# Dialectograms

## Machine Learning Differences between Discursive Communities

Thyge Enggaard[1], August Lohse[1], Morten Axel Pedersen[1,2], and Sune Lehmann[1,3]

[1]Copenhagen Center for Social Data Science, University of Copenhagen, Denmark
[2]Department of Anthropology, University of Copenhagen, Denmark
[3]DTU Compute, Technical University of Denmark, Denmark

## Abstract

Word embeddings provide an unsupervised way to understand differences in word usage between discursive communities. A number of recent papers have focused on identifying words that are used differently by two or more communities. But word embeddings are complex, high-dimensional spaces and a focus on identifying differences only captures a fraction of their richness. Here, we take a step towards leveraging the richness of the full embedding space, by using word embeddings to map out *how* words are used differently. Specifically, we describe the construction of *dialectograms*, an unsupervised way to visually explore the characteristic ways in which each community use a focal word. Based on these dialectograms, we provide a new measure of the degree to which words are used differently that overcomes the tendency for existing measures to pick out low frequent or polysemous words. We apply our methods to explore the discourses of two US political subreddits and show how our methods identify stark affective polarisation of politicians and political entities, differences in the assessment of proper political action as well as disagreement about whether certain issues require political intervention at all.

## 1 Introduction

Over the last decade, the field of Natural Language Processing (NLP) has developed increasingly refined models for numerically representing and quantifying the content of texts. These models serve a variety of purposes, such as classifying or partitioning text, translating texts between languages and engaging in dialogues through bots and APIs. In many cases, these models rely on embedding words as high-dimensional vectors, that represent how words are associated in a training corpus.

For social scientists, these models provide a quantitative way to study how social actors communicate. Word embeddings, in particular, enable researchers to quantify the similarities (or distances) between the use of words among groups of language users. To study how a single word is used, word embeddings can help identify which words are used most similarly in a particular corpus. This approach has e.g. been applied to examine the semantic neighbourhoods of obesity-related words in articles from the New York Times [Arseniev-Koehler and Foster, 2020]. In addition, to study how distinctions are used (e.g. 'woman' vs 'man'), word embeddings can help identify how relatively similar words are to each part of the distinction. Several studies have used this approach, e.g. to study how occupations are associated with the distinctions of gender and ethnicity [Bolukbasi et al., 2016, Caliskan et al., 2017, Garg et al., 2018, Jones et al., 2020]. In these studies, the authors identify a direction in the embedding space, that corresponds to the distinction in question (e.g. 'woman' vs 'man'). By projecting a second set of words (e.g. a set of occupations like 'engineer' and 'secretary') onto this direction, the word embeddings provide a way to measure how these words are associated with the distinction (e.g. 'engineer' being relatively closer to 'man', and 'secretary' relatively closer to 'woman').

Beyond examining how words are related within a single corpus, word embeddings also provide a way of quantitatively identifying and mapping structural differences and similarities *between* corpora. In one strain of research, embeddings of corpora from different languages (like English and Spanish) have been used to automatically translate between the languages [Mikolov et al., 2013b]. In another line of work, word embeddings of corpora from different points in time have been used to detect how the use of words shifts over time [Hamilton et al., 2016, Schlechtweg et al., 2020].

Recently, a third kind of computational comparison has been pursued: the comparison of corpora from two or more communities, who are communicating in the same language (English), during the same period, and roughly about the same topics or events [KhudaBukhsh et al., 2021, Milbauer et al., 2021, Azarbonyad et al., 2017]. While this kind of comparison might be methodologically similar to machine translation and semantic change detection, it provides a way to study the ideological differences between contemporary discourse communities, or what some have dubbed 'ideolects', "a language shared by an ideological group, which necessarily contains the private worldview of that group and may not be naively decipherable to those outside" [Milbauer et al., 2021].

Until now, such work has mainly focused on *identifying* which words are used differently, typically by either measuring the distance between the embeddings of each word or by identifying words, that mistranslate between the embeddings.[1] While such mistranslations will include contingent spelling errors and near-synonyms ('lmao', 'lol'), they also detect structural equivalences between the communities, such as the similarity between the use of 'kkk' and 'blm' in the commentaries of CNN and Fox on YouTube, or between 'son' and 'daughter' in the subreddits r/askmen and r/askwomen [KhudaBukhsh et al., 2021, Milbauer et al., 2021].

But what do these differences *consist* of? While previous work offers examples of the semantic contexts that might explain the mistranslations (such as kkk/blm as 'a hate group' in the commentary to CNN/Fox), no systematic way of accounting for the difference in embedding has so far been proposed.

Here, we propose a method for calculating and visualizing the difference in how discursive communities use a given word. We refer to these visualisations as *dialectograms*. More precisely, we show how the projection of words along the difference in the embedding of a focal word captures the characteristic ways in which each corpus uses the focal word.

In addition, we also show how dialectograms allow us to construct a new measure of difference in word use, that overcomes the tendency for cosine distance and mistranslations to pick out low-frequent and polysemous words (words with multiple senses, like 'settlement'). By calculating the degree to which the characteristic use of each corpus is also *unique* to each corpus, our measure of *sense separation* is designed to single out words, which each corpus use in rather distinct ways.

In the following, we describe and illustrate our method by comparing two US political subreddits, r/Democrat and r/Republicans, covering the period from January 2011 to September 2022. We apply a range of preprocessing steps, including lemmatisation and restricting both vocabularies to words that appear at least 100 times in both corpora. After preprocessing, the r/Democrats corpus consists of 511.906 comments and the r/Republican of 576.872 comments. For a detailed description of our data collection, preprocessing and corpora construction, see the Materials section.

We begin by providing an example of a dialectogram, after which we describe how it is constructed. We then describe *sense separation*, our new measure of difference in word use. Finally, we use sense separation in concert with the dialectograms to carry out a detailed, exploratory analysis of our two corpora. This analysis reveals a stark affective polarisation of partisan references, mainly characterised by derogatory cross-partisan discourse. While partisan references exhibit affective polarisation, there is, however, also a range of non-partisan references exhibiting high difference in use. In the case of the non-partisan references, the dialectograms suggest that the differences rather arise from divergent assessments of which policies to pursue, as well as whether certain topics require political engagement at all. Finally, we show that on aggregate across all words, the characteristic use in r/Democrats tends to focus on electoral and legislative issues, while the characteristic use in r/Republican tends to focus on (contentious) topics of policy.

## 2 Results

To begin, Figure 1 shows the dialectogram for the word *republican*.[2] The words associated with the characteristic way *republican* is used in r/Republican are located in the upper right quadrant; they centre around *candidates* and how they are perceived (e.g. *misunderstood*, *constructive*, *hopeful*). In addition, there is a range of words related to Reddit itself (e.g. *post*, *sub*, *comment*), that stem from internal talk about the republican subreddit[3]. Opposed to this is the characteristic use of *republican* in r/Democrats, located in the lower left quadrant, which appears to be mainly derogatory (e.g. *filth*, *treasonous*, *brainwash*).

[1] A word mistranslates from one embedding to another, if its nearest neighbour in the other embedding is not the word itself.

[2] One criticism of unsupervised text analysis, like we deploy here, is the reliance on a relatively low number of words to interpret the fitted models. For this reason, we provide full-page visualisations, with a large number of annotated words. While the annotations might occasionally clutter, we believe they overall provide a more detailed yet still readable way to interpret the dialectograms.

[3] We conclude this based on sampling and reading comments in the r/Republican subreddit, that include these words

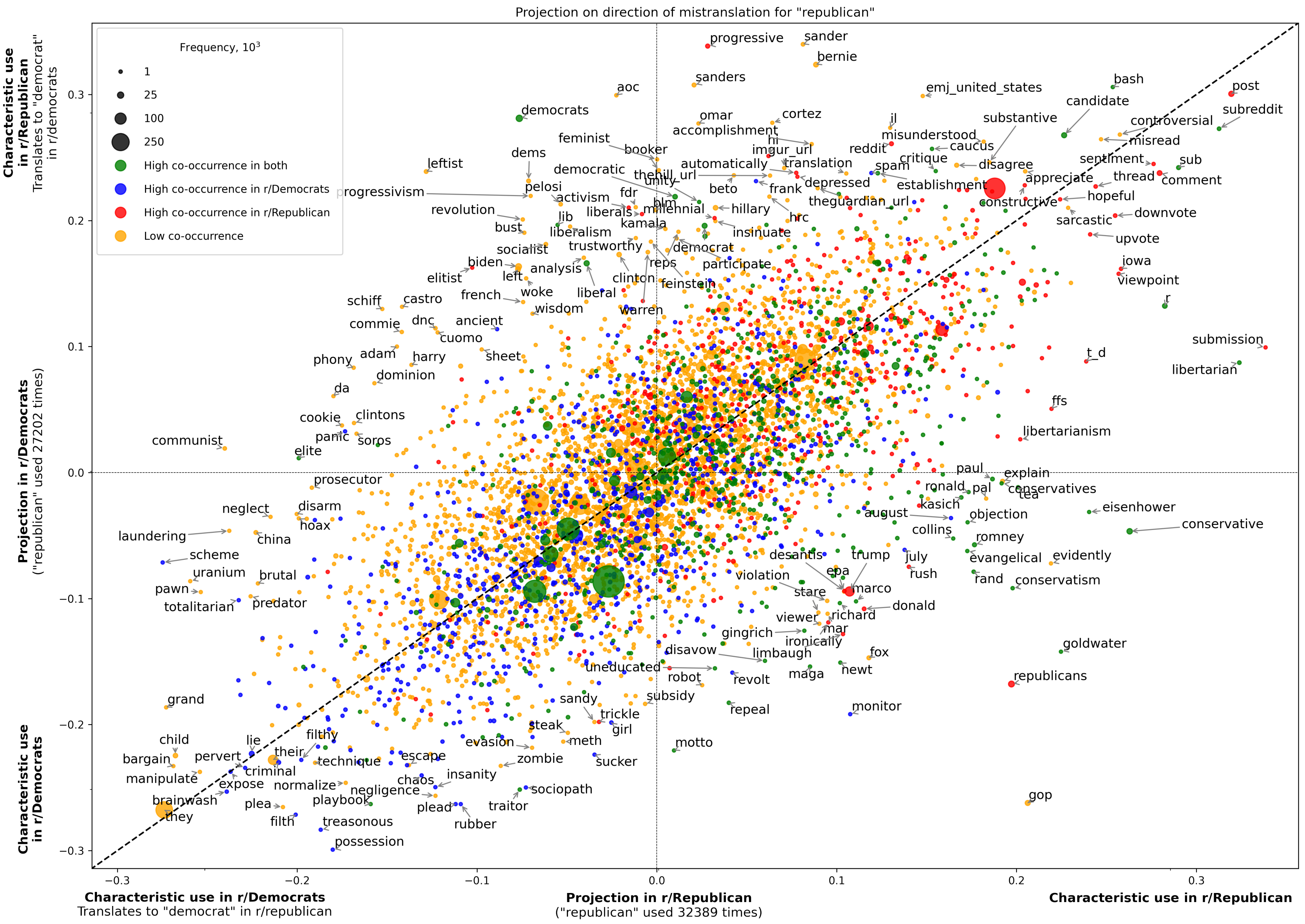

Figure 1: Projection of words on the offset for the embeddings of *republican*. The dispersion of words along the x-axis might appear greater than along the y-axis, as the figure has been stretched to the page. Words are coloured according to their co-occurrence with *republican*; see Equation 1 for the definition of high co-occurrence.

The words that both communities associate more with their own use of *republican* are located in the lower right quadrant. As can be seen, these tend to be more descriptive in the form of near-synonyms (*republicans*, *gop*), names of republican politicians (e.g. *goldwater*, *donald*), movements (*maga*, *tea*), or *conservative* and *evangelical* associations. Words that each community associate more with the other community's characteristic use are located in the upper left quadrant. Again, we see that these words tend to be descriptive, this time in the form of references to democrats (e.g. *democrats*, *leftist*, *pelosi*). Inspecting the translation of *republican* between the corpora (provided on the axis labels in Figure 1), this is due to the strong discursive equivalence of *republican* and *democrat* in these subreddits; the way r/Republican uses *republican* is most similar to the way r/Democrats uses *democrat*, while the way r/Democrats uses the word *republican* is most similar to the way r/Republican use *democrat*.[4]

We will return to a more extensive analysis of the partisan divide in section 2.3 below. But first, we describe how the dialectogram in Figure 1 is constructed (section 2.1), as well as how dialectograms enable us to suggest a new measure of difference in use (section 2.2).

## 2.1 Constructing dialectograms

Constructing the dialectogram in Figure 1 involves three overall steps. First, we embed each corpus, which requires choosing an embedding model. Second, we align the embeddings, such that the embedding of words from different corpora can be compared. This requires choosing an alignment method. Third, to create the dialectogram for a focal word, we project each embedding onto the direction corresponding to the difference in the embeddings of the focal word. For each word, we obtain two projections, corresponding to the projection from each of the two corpora. We then construct the dialectogram by plotting these scalar projections against each other. Below, we consider each of the steps in turn.

### Embedding the corpora

In an exploratory analysis like ours, there is no straightforward way to decide which embedding model is best, since we do not have a direct way to compare their performance. We therefore devise a validation task, that evaluates how well different embedding models are able to recover the degree to which words are used differently between two validation corpora. To control the degree to which words are used differently, we create a modified version of the r/Republican corpora, where we pair and swap $\sim 11\%$ of words. By varying the degree to which we swap words within each pair, we are able to compare how well the different embeddings recover the degree of change based on a range of different measures.

In the Model and Measure Selection section, we describe the validation task further and provide the results of comparing two static embedding models (the skip-gram with negative sampling (SGNS) version of Word2Vec and GloVe) and one contextual embedding model (DistilBERT, a distilled version of a pre-trained RoBERTa model) [Mikolov et al., 2013a, Pennington et al., 2014, Liu et al., 2019, Sanh et al., 2020].

Overall, we find little difference in how well the three embedding models recover the degree of swap. Notably though, while translating between the static embeddings also recover which word a given word has been swapped with, this is not the case for the centroids from the DistilBERT model. As GloVe embeddings are more stable than SGNS embeddings, we present our results based on GloVe embeddings here, but our method is not restricted to this choice of model [Wendlandt et al., 2018, Antoniak and Mimno, 2018]. For more about how we train the GloVe embeddings, see the Methods section.

### Aligning the embeddings

Training two GloVe embeddings on the same corpora will not yield the exact same numerical structure. Part of this is due to the random initialisation of vectors, providing different initial configurations of the embedding spaces. However, even if the embeddings are not numerically identical, they might still reflect the same structural representation of the corpus in the sense that all pairwise distances between words are the same.

To make the embeddings comparable, we align the embedding spaces by identifying a linear transformation between them. As part of our validation task, we compare two approaches to identifying such linear transformations, Procrustes Analysis and Canonical Correlation Analysis, and find they work equally well [Schönemann, 1966, Artetxe et al., 2016, Balbi and Misuraca, 2006, William, 2011]. As Procrustes Analysis has the desirable property of leaving all pairwise distances within each embedding space unaffected, we use it here. For a description of the two alignment methods, see the Methods section; for the comparison in the validation task, see the Model and Measure Selection section.

[4] To translate a focal word from one corpus to the other, we find the word in the later embedding that is closest to the embedding of the focal word in the former. For more information, see the Methods section.

### Comparing the projections

Having obtained two aligned embedding spaces, we can now create the dialectogram of a focal word i. First, we identify the vector offset between the embeddings of word i, $O_i = w_{1,i} - w_{2,i} \in \mathbb{R}^D$, where $w_{k,i} \in \mathbb{R}^D$ is the embedding of word i in corpus $k \in \{1, 2\}$. Intuitively, the vector offset $O_i$ captures the way word i is used differently in the two corpora, similar to how a vector offset between feminine and masculine words should capture variation relevant to the distinction of binary genders in a corpus. While there are likely more similarities than differences between how a word is used in the two corpora, this approach allows us to home in on the difference, no matter how small.

To measure how the remaining words are related to the difference in use of the focal word, we project the embeddings onto this vector offset (for an illustration, see Figure 2a). Specifically, we obtain the scalar projections from the embedding of corpus k, $E_k \in \mathbb{R}^{N \times D}$ as

$$\alpha_i^k = \frac{E_k O_i}{\| O_i \|_2} \in \mathbb{R}^N$$

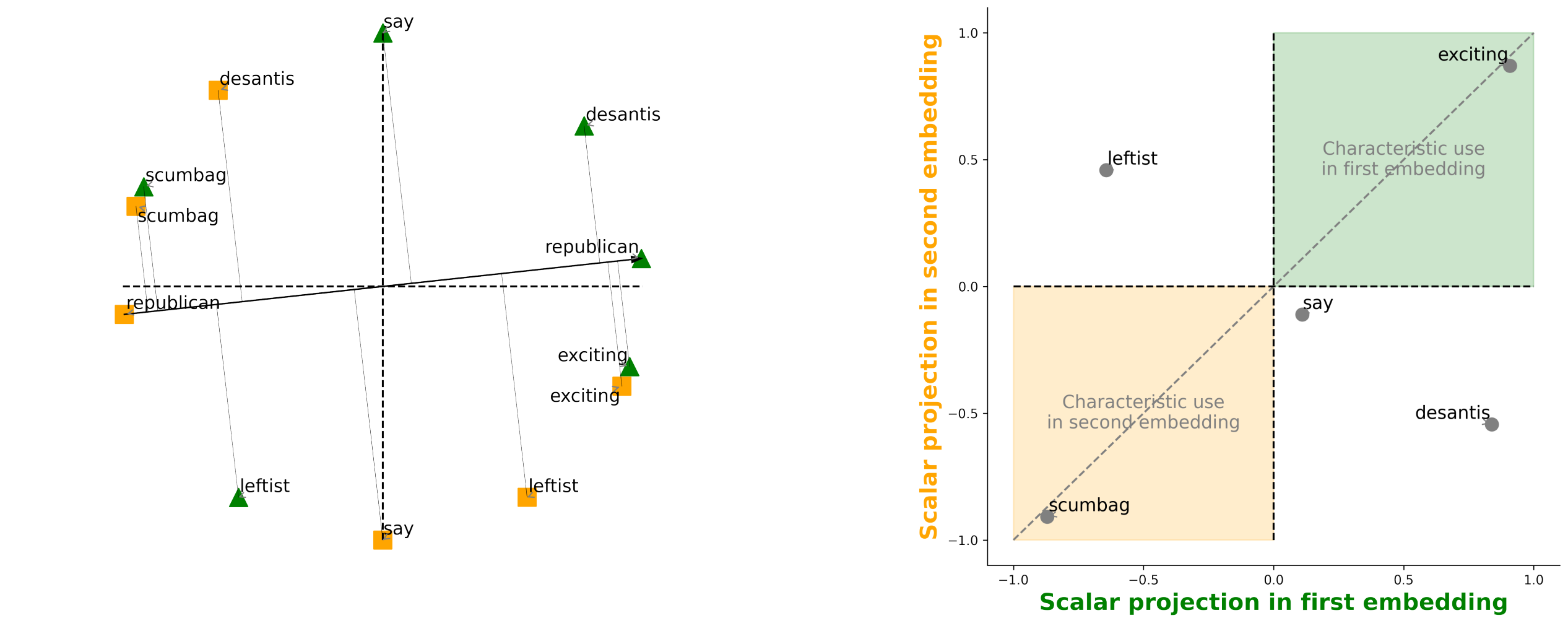

(a) **Projecting onto the offset:** Given two aligned embeddings (one marked with green triangles, the other with orange squares) and a focal word (here 'republican'), we first identify the vector corresponding to the difference between the embeddings of the focal word in the two corpora, marked as the solid black line. We then project the remaining vocabulary onto this offset, to create a graded measure of how each corpus relatively associates each word with the two embeddings of the focal word.

(b) **Comparison of scalar projections:** Plotting the two scalar projections of each word against each other, we can distinguish between the characteristic ways each corpus associates the focal word ('exciting' vs. 'scumbag', areas marked with green and orange respectively). In addition, words in the two remaining quadrants ('desantis', 'leftist') are relevant to the distinction between the characteristic uses, but the corpora do not agree to which characteristic use they belong. Finally, words close to the centre are not relevant to the difference in use.

Figure 2: Illustration of our geometric approach to accounting for the disputed conceptions of a word

By plotting the scalar projections from each embedding against each other, we can distinguish and interpret the different ways in which the two corpora use the focal word (see Figure 2b for an illustration). Overall, we distinguish between three scenarios.

First, words projecting relatively close to zero in both embeddings ('say' in the example in Figure 2b) are discursively unrelated to the difference in how the focal word is used. This can either be because they are unrelated to the focal word in both corpora, or because their relation to the focal word is similar in both corpora.

Second, words projecting positively in both embeddings ('exciting', upper right quadrant marked with green) are, in both aligned embeddings, closer to the embedding of the focal word in the first (green) embedding, and hence account for the characteristic way in which the first corpus uses the word. Similarly, words projecting negatively in both embeddings ('scumbag', lower left quadrant marked with orange) account for the characteristic way in which the second corpus uses the word. In general, these make up the *relative* difference in how the corpora use the focal word ('republican') - both corpora might still use the word in both ways, just to varying degrees.

Third, words projecting positively in the first embedding but negatively in the second ('desantis', lower right quadrant) are more associated with each corpus' own characteristic use. Likewise, words projecting negatively in the first embedding but positively in the second ('leftist', upper left quadrant) are, in each corpus, more associated with the other corpus' characteristic use. Both corpora associate the words in these quadrants with one of the characteristic uses but do not agree to which characteristic use they belong.

In practice, we find that only plotting the projections of words, that co-occur at least once with the focal word in either of the corpora, makes the dialectograms easier to interpret. While the projection of non-co-occurring words might provide a way to assess the most implicit associations within each corpus, we found they muddled our interpretation. Further, we leave out the three highest frequent words (*be*, *do*, *have*) in order to illustrate word frequency by marker size, as well as the focal word itself.

## 2.2 Identifying words used most differently

A dialectogram can be constructed for any word that appears in both of the aligned embeddings. As our vocabulary consists of 5.738 unique words, it is infeasible to inspect the dialectograms of all words. If a researcher has a particular theoretical or empirical emphasis, this might provide a way to select which words to create dialectograms of. Here, we instead use methods for identifying words that appear to be used most differently between two corpora.

To identify such differences, we need a measure of the degree to which words are embedded differently. Several measures of difference between embeddings exist. As part of our validation in the Model and Measure Selection section, we describe and compare some of these. In general, we find cosine distance best recovers the degree of swap. Further, we find that whether a word mistranslates or not is highly correlated with whether the word is swapped more or less than 50%. In the following section, we present a new measure of difference in use referred to as *sense separation*. Sense separation is based on the dialectograms and is designed to overcome the tendency for cosine distance and mistranslation to pick out low-frequent or polysemous words in our application. We briefly cover these challenges, before describing the sense separation measure.

#### Difference in use and frequency

Figure 3a shows the relation between aligned cosine distance and the log of mean frequency across the two corpora. Cosine distance correlates negatively with frequency (Spearman's correlation: -0.51), implying that words exhibiting high cosine distance tend to be used less. As a larger distance makes translation errors more likely, mistranslations are also more prevalent among low-frequent words. This issue can partially be overcome by focusing on the mistranslating words, that have the highest frequency.

In the case of semantic change detection, this negative correlation has been interpreted as a substantial feature of language [Hamilton et al., 2016]. In our comparative setting, it seems unlikely, however, that the degree to which a word is used differently between two corpora should be determined by the frequency of the word. Instead, it is possible that the negative correlation at hand is a property of the embedding models themselves, an explanation supported by controlled experiment [Dubossarsky et al., 2017]. In the Methods section we describe this challenge further, along with our (failed) attempts to remove the correlation.

One adjustment does however support the interpretation of the dialectograms: projecting each embedding to the orthogonal complement of the direction that varies most with the logarithm of word frequency. Intuitively, projecting the embeddings to the complement of this direction should remove the information about frequency implicitly embedded in the vector spaces.

Unfortunately, this strategy does not remove the negative correlation, which suggests that the relation between frequency and embedding position is not encoded in a single direction. However, for words that are used much more by one corpus than the other, this adjustment does balance the projections in the dialectograms. Without this adjustment, the characteristic use in the high-frequent corpus tends to be characterised by a small tail of high-frequent, often generic words - after the adjustment, the projections are balanced around the origin, and the corresponding dialectograms become more clear to interpret. For this reason, we apply the adjustment to the embeddings, before we align them (for a description of this adjustment, see the Methods section).

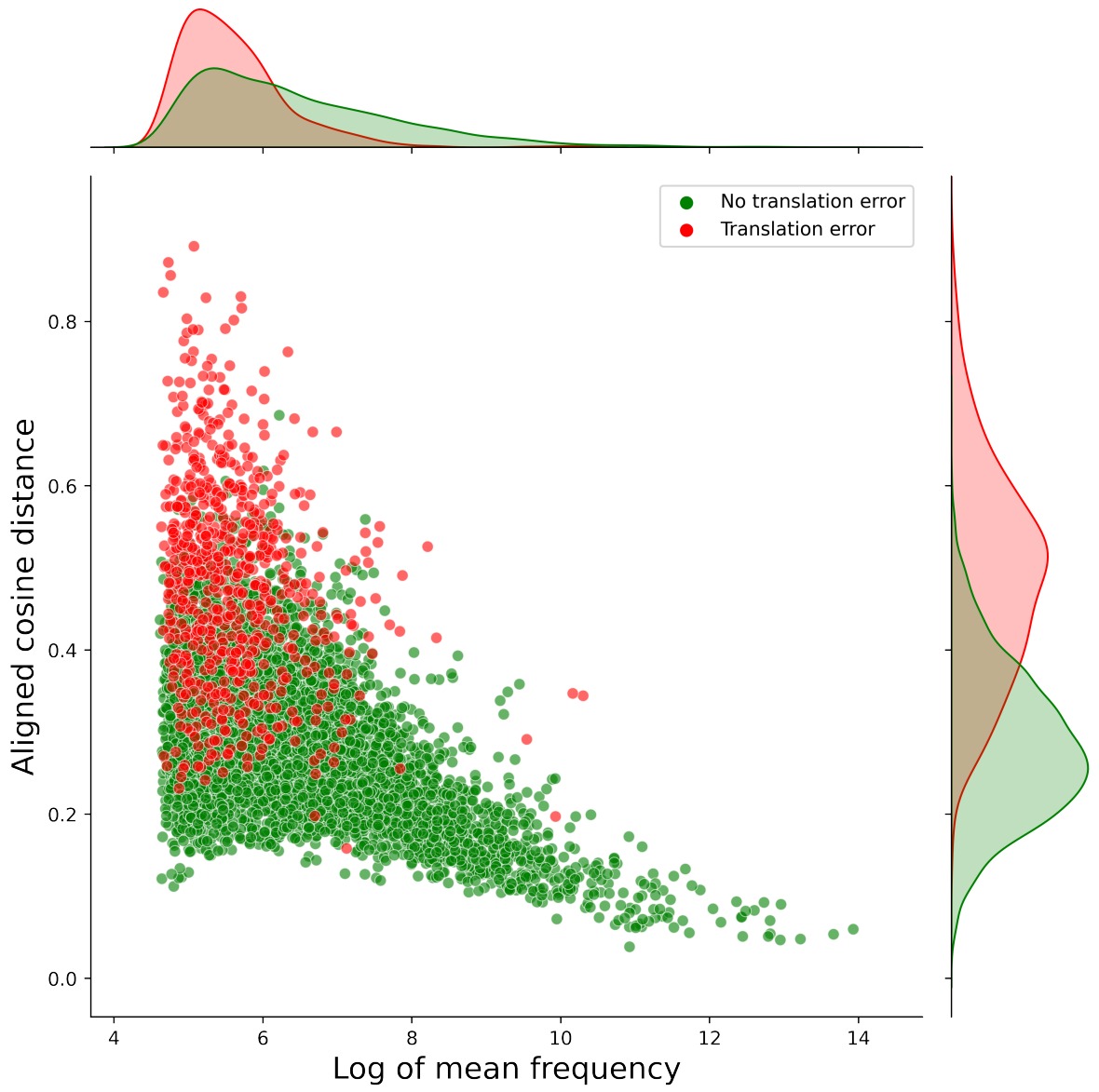

(a) **Cosine distance:** Scatter plot of the log mean frequency of each word against its aligned cosine distance. Cosine distance correlates negatively with frequency, why it tends to identify low-frequent words as used most differently.

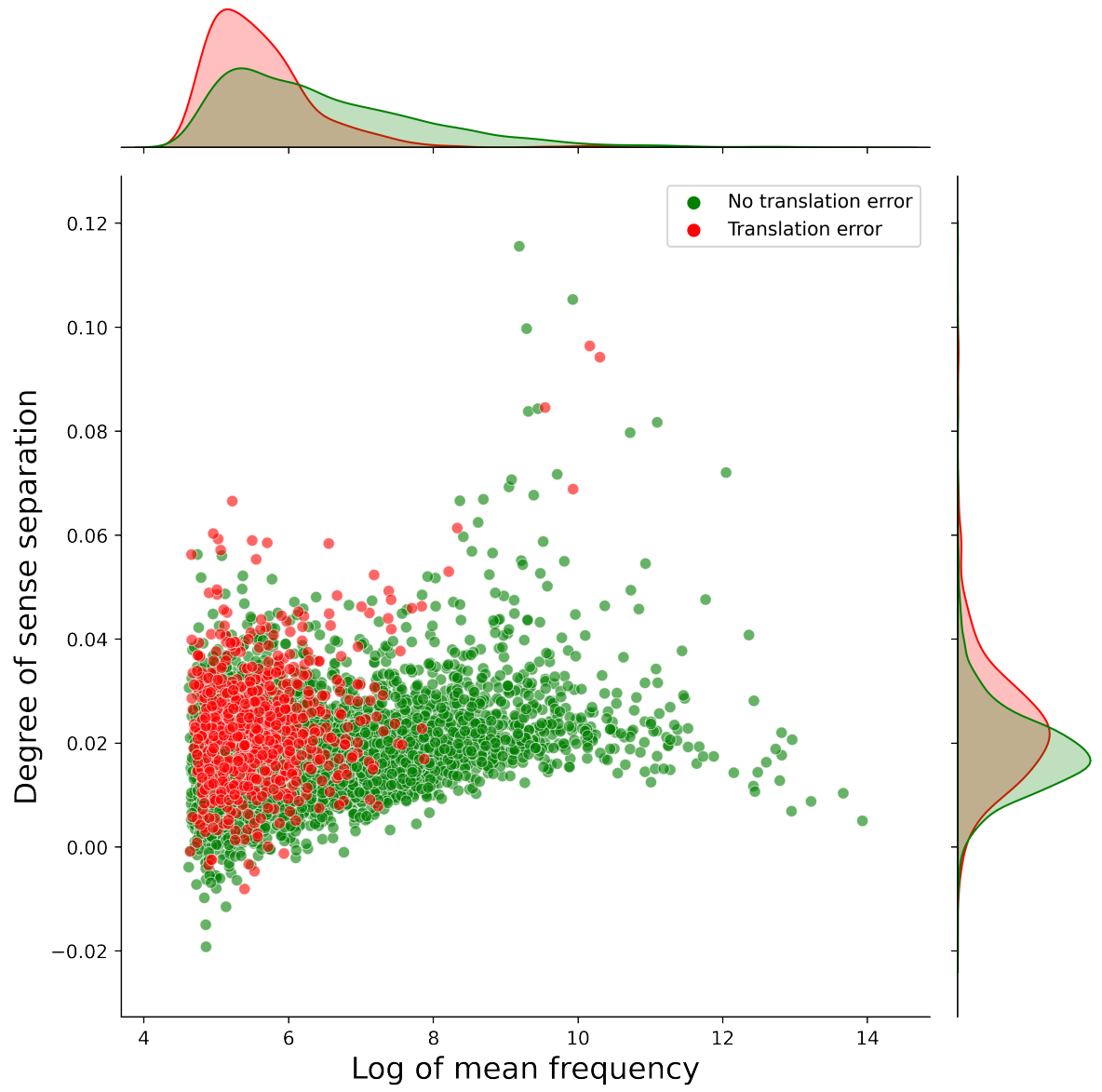

(b) **Sense separation:** Scatter plot of the log mean frequency of each word against its degree of sense separation. Sense separation identifies a range of higher-frequent words as used most differently; these also include the highest-frequent mistranslations.

Figure 3: **Relation between measures of difference in use and frequency**. Words are coloured according to whether their two embeddings are each others' nearest neighbours across embeddings (no translation error) or not.

### Difference in use and polysemy

In addition to correlating with frequency, cosine distance and high-frequent mistranslations also tend to pick out polysemous words when applied to our corpora. The word *settlement* e.g. has a high cosine distance; by inspecting the dialectogram, the two characteristic uses correspond to the sense of an inhabited place vs. an agreement. The dialectogram also indicates that both corpora use the word in both senses, as evidenced by the presence of words in both of the characteristic uses, that co-occur highly with *settlement* in both of the corpora. In general, even if the polysemous word is used in the same set of senses by both corpora, differences in the degrees to which each sense is used result in differences in the embedding of the word [Arora et al., 2018].

In some cases, such differences might be particularly interesting for social scientists. For example, we can see that r/Republican tends to talk relatively more about a metaphorical *flood* of illegal immigration, while r/Democrats are relatively more occupied with the management of a fluid *flood*. But there are also cases, where the difference in the degree to which the senses of a polysemous word are used, might be less interesting. The word *sheet* e.g. exhibits high cosine distance - inspecting the dialectogram, we interpret this as the difference between a *sheet* of toilet paper and a fact *sheet*, such as published by institutions like the White House.

### Sense Separation

Based on the dialectograms, we can now devise a measure of the degree to which the characteristic use of each corpus is also unique to that corpus. In the case of polysemous words, one would intuitively expect this measure to be high in the cases where the multiple senses are split and not shared between the corpora (e.g. one corpus writing solely about physical settlements, the other solely about legal agreements), which we consider unlikely to occur.

The intuition behind sense separation is illustrated in Figure 4. Here, we show two hypothetical dialectograms, that correspond to a low and high degree of sense separation. In these hypothetical dialectograms, the projected words are coloured according to whether they co-occur highly with the focal word in one or the other corpus. Our sense separation measure captures the degree to which these highly co-occurring words are separated between the two characteristic uses. In particular, Figure 4a illustrates a case where there is almost no separation between the uniquely high co-occurring words (red and blue dots); in this case, the sense separation measure will be low.

In contrast, Figure 4b illustrates a case, where the separation is much more pronounced; in this case, the sense separation measure is high. For a definition of the sense separation measure, see the Methods section.

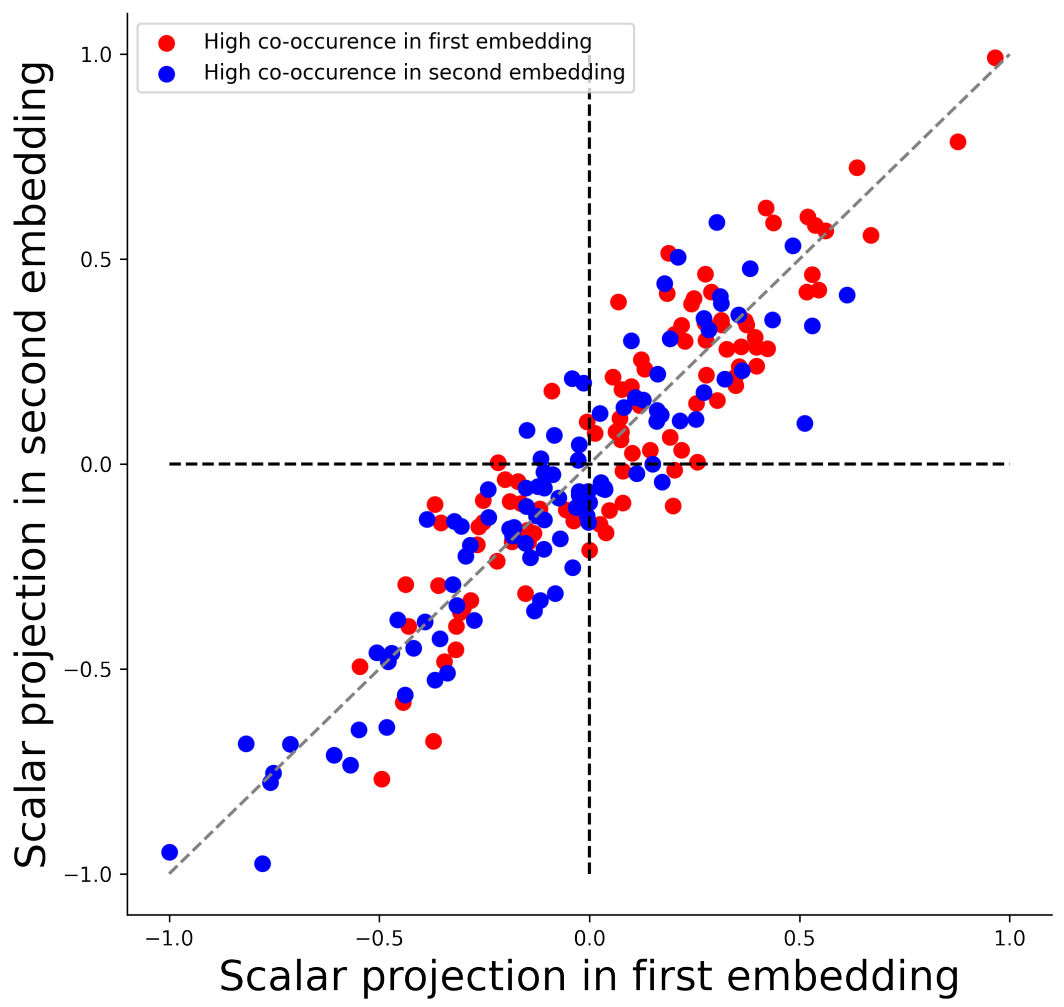

(a) **Low sense separation:** The words that co-occur highly with the focal word in one or the other corpora (red and blue dots) are not separated between the upper right and lower left quadrant.

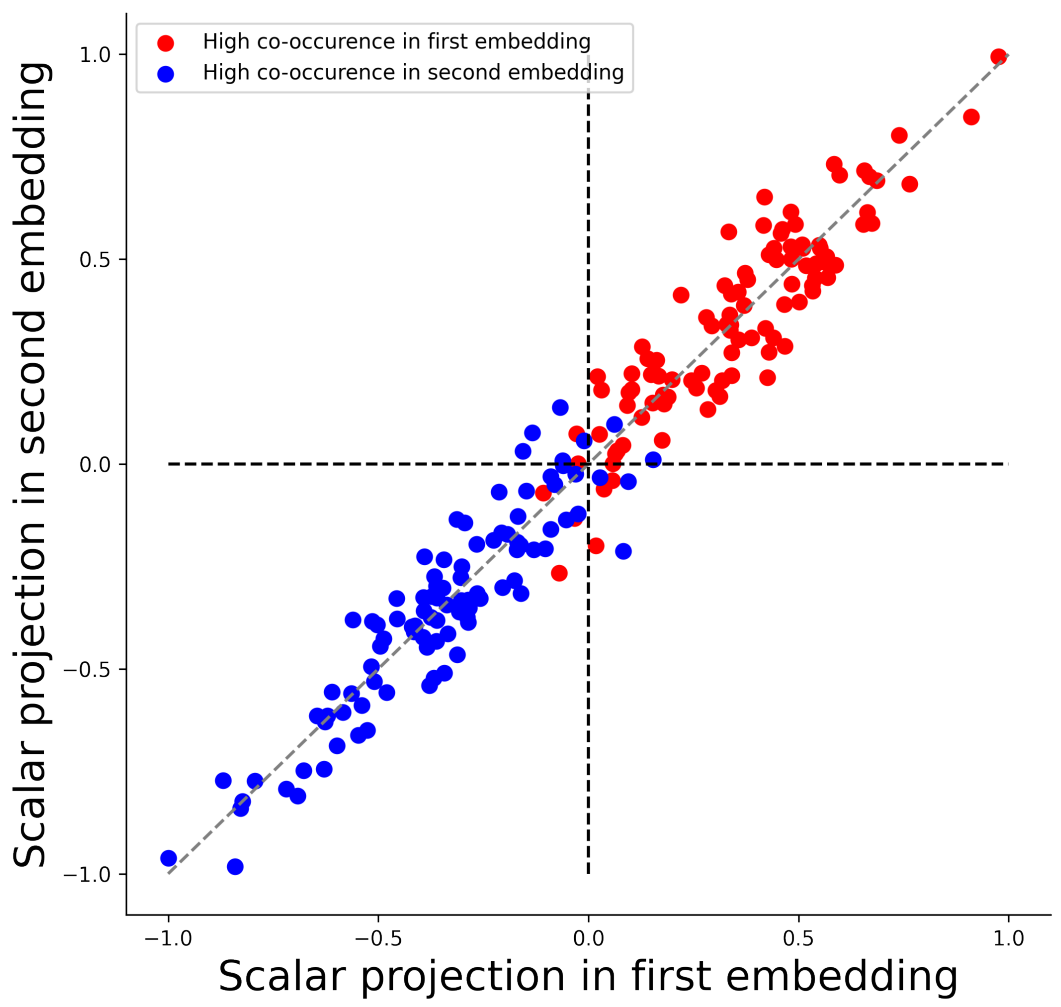

(b) **High sense separation:** The words that co-occur highly with the focal word in one or the other corpora (red and blue dots) are much more separated between the upper right and lower left quadrant.

Figure 4: **Illustration of the intuition behind sense separation**: Our measure of difference captures the degree to which the words, that co-occur highly with the focal word in only one of the corpora, are separated between the two characteristic uses.

As Figure 3b shows, the sense separation measure is less correlated with frequency than cosine distance (Spearman's correlation: 0.16). Further, the highest-frequent mistranslations also show a high degree of sense separation.

### 2.3 Comparing r/Republican and r/Democrats

Having described the construction of the dialectograms, as well as our measure of sense separation, we now return to a more detailed analysis of the two subreddits.

Table 1 shows the 25 words used most differently in our two corpora, based on (i) cosine distance, (ii) highest frequent mistranslations and (iii) sense separation (see the Supplementary Information for top 100 words). Inspecting the lists, only six words appear in more than one (*republican*, *republicans*, *democrats*, *liberal*, *newt* and *cruz*), and none in all three, highlighting that the three measures capture distinct differences between the embeddings. Further, several words are clearly partisan, such as names of politicians (e.g. *biden*, *cruz*, *gingrich*) and party references (e.g. *republicans*, *democrats*, *gop*). We also see references to ideologies (*liberal*, *conservative*), concrete political issues (the *keystone* pipeline) and the media (*forbes_url*, and *pbs*, the Public Broadcasting Service).

There is, however, also a range of words, that does not seem to fall within any of the categories mentioned above, suggesting that the discursive differences extend beyond these somewhat expected categories. These words include the emojis, of which several have to do with laughing (😂) and mocking (🤡); various references to the internet and Reddit itself (*subreddit*, *r*, *com*, *url*); as well various functional words, including stop words that are often ignored in content analysis (e.g. *lastly*, *unfortunately*, *she*, *he*).

We will now examine some of these differences by using our dialectograms. We first turn to the partisan references, before proceeding to some of the non-partisan words. Finally, we consider the aggregate differences between the corpora's characteristic uses. While we only show the dialectograms for a few words here, the dialectograms for all words in Table 1 can be found at https://github.com/A-Lohse/Dialectograms.

| Highest cosine distance | Highest-frequent mistranslations | Highest sense separation |
|---|---|---|
| voluntary | republican | r |
| juvenile | democrats | biden |
| com | republicans | comment |
| 🤪 | liberal | democrats |
| gingrich | cruz | republican |
| sheet | leftist | liberal |
| 👋 | unfortunately | gop |
| settlement | subreddit | left |
| keystone | sander | trump |
| newt | warren | she |
| spur | 😂 | he |
| ❤️ | kasich | conservative |
| karl | yep | hillary |
| airline | bs | clinton |
| captain | harris | republicans |
| mission | desantis | democrat |
| port | orange | race |
| rescue | rubio | sanders |
| rove | approve | pbs |
| cough | son | progressive |
| 🤡 | 🤣 | cruz |
| 🇺🇸 | mitch | obese |
| url | ted | fox |
| lastly | rand | broadband |
| forbes_url | ah | newt |

Table 1: Words used most differently, as measured in three ways

## Partisan Discourse

As Figure 1 shows, the difference in how the corpora use *republican* revolves around within-party candidate evaluation versus across-party derogatory talk. Inspecting the dialectograms of other partisan references and politicians, this pattern consistently re-appears, indicating a general discourse for evaluating partisan affiliations. To illustrate this pattern, Figure 5 projects US politicians along the mean direction of their offsets [5,6]. To understand what discursive variation the mean offset corresponds to, we also plot the words that overall project most negatively or positively on the mean offset in each embedding.

Overall, we see the same opposition between derogatory words (lower left quadrant) versus evaluating and appraising candidates (upper right quadrant) reappear. This pattern suggests that there is a strong degree of affective political polarisation between our corpora [Iyengar et al., 2012]. We further note that cross-party animosity seems particularly strong as compared to within-party appraisal. While the positive associations remain within the setting of political candidates (e.g. *substantive*, *pragmatic*, *enthusiasm*), the negative associations do not seem bound to an electoral or legislative domain but are outright derogatory (*filth*, *bastard*, *rapist*). The particular salience of out-group discourse is also suggested in recent research on political polarisation, which finds posting about political out-groups to be the strongest predictor of whether a social media post is shared [Rathje et al., 2021].

In addition, the relative positioning of politicians in Figure 5 seems to suggest that both communities are somewhat left-leaning within their partisan affiliation; while politicians like *trump*, *gaetz* and *mcconnell* projects fairly neutral in r/Republican, *mitt romney* projects clearly toward the positive end of the spectrum; and likewise, while *nancy pelosi* projects neutral in r/Democrats, politicians like *elizabeth warren* and *bernie sanders* project positively.[7]. We also see that there are some differences between the words the two corpora associate with each end of the spectrum - notions of *commie* and *communist* e.g. project negatively in r/Republican but neutral in r/Democrats, that in turn seems to associate the upper right quadrant more with *activism* and *progressivism*.

[5] Any shared distinction is not unique in the direction of the offsets - in the embedding of r/Republican, we e.g. expect derogatory words to project negatively for republican politicians, but positively for democratic politicians. Hence we calculate the mean offset across all politicians' offsets (Republicans and Democrats), calculate the cosine similarity between each politician's offset and this mean offset, flip the direction of the offset for all politicians, whose cosine similarity to the mean is negative, and finally re-calculate the mean offset, before we project all politicians onto it.

[6] We identify politicians referenced in the corpora by inspecting words tagged as proper nouns

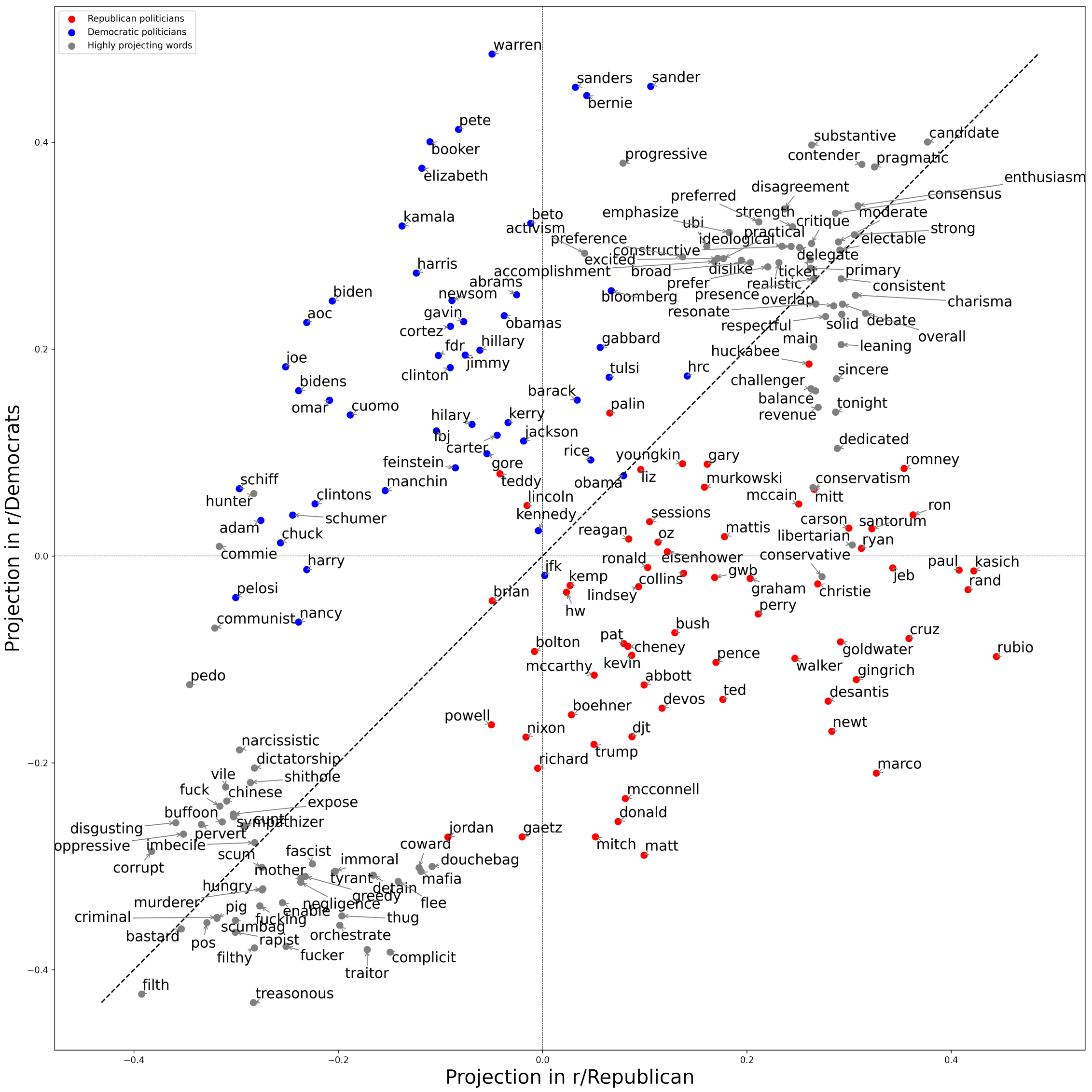

Figure 5: Projection of words on the mean offset of US politicians.

While explicit partisan references like politicians exhibit this kind of affective polarisation, the projection on *republican* in Figure 1 indicates that this also extends to third-person-pronouns (*they*, *their*). Unsurprisingly, given that these likely serve as indirect partisan references, the singular third-person pronouns that appear in Table 1 (*she*, *he*) as well as their possessive counterparts a little further down the list (*her*, *his*) display a further trend; while derogatory words are particularly associated with both *she* and *her* in r/Republican, they are reversely associated with *he* and *his* in r/Democrats. This could be reflective of a greater share of prominent women in the Democrats party - inspecting the dialectogram for *she*, the dispute seems to be particularly pronounced in the cases of Alexandria Ocasio-Cortez, Ilhan Omar and Stacey Abrams. In addition to this gendered opposition between the corpora, we also note that the derogatory talk seems different in the case of *he*. This seems to suggest that while derogatory talk about *her* acts and attributes go together with derogatory talk about who *she* is (cosine similarity between offsets is 0.56), this link is less clear in the case of *his* and *he* (cosine similarity is 0.17).

[7]We are cautious in interpreting the relatively *shared* positive projection of *huckabee*, *palin* and *sanders*, as inspecting the dialectograms of the names highlights a drawback with our method; as Sarah Huckabee Sanders shares parts of her name with both Sarah Palin and Bernie Sanders, the associations of these names are likely entangled.

### Non-partisan discourse

While the explicit partisan references exhibit affective polarisation, such polarisation is less evident, if appearing at all, when we inspect the dialectograms of the non-partisan words in Table 1. For example, Figure 6 shows the projection on the offset for *juvenile*. In this case, two different conceptions of *juvenile* seem to be at play. On one hand, the use in r/Republican is characterised by notions such as *violent*, *insulting* and *disrespectful*, calling for *arrest* and *incarcerate*. In addition, the *population* is salient, particularly with regard to race and ethnicity (*latino*, *black*, *white*, *hispanic*). Contrary to this, the use of *juvenile* in r/Democrats is characterised by a focus on *reform*, *justice*, *system*, with greater emphasis on the *defender* and the financial implications (*bail*, *economic*). While both corpora associate their own characteristic use with notions like *abusive* and *embarrassing*, they associate the other corpus' use more with *ego*.

Here, then, the difference does not appear affective, but rather policy-oriented. Other topical words also exhibit non-affective differences. The *keystone* pipeline is e.g. related to an industrial discourse in r/Republican (e.g. *gas*, *oil*, *production*, *drive*) but with macro-economics words (*unemployment*, *gdp*, *benefit*) and a focus on *jail* and *sentence* in r/Democrats. Among the top 100 words (see Supplementary Information), a similar case is *port*, with industrial associations in r/Republican and more general notions of *infrastructure*, *investment* and *prosperity* for *america* in r/Democrats. Examples also include *competitor* (more domestic in r/Republican, more international in r/Democrats) and *homelessness* (associated with *neighborhood crime* in r/Republican and broader set of words like *China*, *climate* and *veterans* in r/Democrats).

In addition to these rather clear cases of partisan affective polarisation and differences in topical perspective, we finally observe cases, where the differences in how words are used seem more related to assessments of veracity and political relevance. This point is perhaps most clearly illustrated by the words *greenhouse* and *vaccination*. While the characteristic use of these words in r/Republican seems to convey an overall sceptical message, through associations like *effect* and *outcome* together with *science*, *statistic*, *proof* and *(dis)prove*, the characteristic use of *greenhouse* and *vaccination* in r/Democrats seem to be more associated with policy issues and matters of governance.

### Overall differences in the characteristic use of words

Let us present one final example of what kind of pattern detection and social scientific analysis that our method opens up for. While inspecting the dialectograms of the words identified in Table 1, we noticed another pattern between the characteristic uses. Across several dialectograms, the characteristic use in r/Democrats tends to focus on electoral entities and processes, whereas the use in r/Republican tends to be relatively more characterised by (contentious) topics of politics. This difference in emphasis is in line with the subreddits' self-description; while r/Republican is presented as "a partisan subreddit [..] a place for Republicans to discuss issues with other Republicans", r/Democrats is conversely introduced as a subreddit that "offers daily news updates, policy analysis, links, and opportunities to participate in the political process. We are here to get Democrats elected up and down the ballot." [r/Republican, 2022, r/Democrats, 2022]

To aggregate the overall differences between the characteristic uses in the two corpora, we count how often the mean projection of each word across all offsets is greater than 0.2 and less than -0.2, equivalent to words that most often project heavily towards the characteristic use in r/Republican (upper right quadrant) or in r/Democrats (lower left quadrant). Then we subtract the two counts, focusing on words that tend to project heavily towards the characteristic use in either corpus, but not both (such as the partisan associations identified above).

Table 2 shows the 30 words with the highest and lowest value, corresponding to words that tend to project heavily towards the characteristic use in either r/Republican and r/Democrats, respectively. This confirm our observation from the individual dialectograms; while electoral (e.g. *election*, *voter*, *lose*, *seat*), legislative (*policy*, *majority*, *governor*), and judicial words ( *court*, *judge*, *scotus*) appear frequently in the list for r/Democrats, words that most unanimously project towards the use in r/Republican include words related to *woman* (*marriage*, *fetus*), immigration (*immigrant*, *immigration*), healthcare (*disease*, *vaccinate*, *covid*), policing (*police*, *officer*), macroeconomics (*spending*, *budget*, *dollar*) as well as notions of *freedom*, *government* and *corrupt*. A word that illustrates this general difference is *suppression* (Figure 7). Here, notions of *legislature* and *gerrymandering* are salient in r/Democrats, while r/Republican seem more occupied with the suppression of *speech* and *information* through *censorship*.

## 3 Conclusion

The proliferation of social media and other digital infrastructures provides an unprecedented archive to study how language is used by different individuals and groups. With large amounts of text generated every day, for prolonged periods, social scientists are not only able to study how words are used in a single corpus, but also to compare and contrast how words are used across time and space. One way to computationally do so is to embed the co-occurrences of words into geometric spaces, which can subsequently be aligned and contrasted. By encoding the totality of relations between all pairs of words within a given corpus, this method can map distinctions of discourse into the structure of the geometric spaces [Mikolov et al., 2013c, Kozlowski et al., 2019, Grand et al., 2022]. Such mappings of the full range of word associations, however, make it hard to discover the specific and more fine-grained differences in word usage that are often of social scientific interest - e.g. how Republican and Democratic supporters write differently about the current President of the United States, Joe Biden.

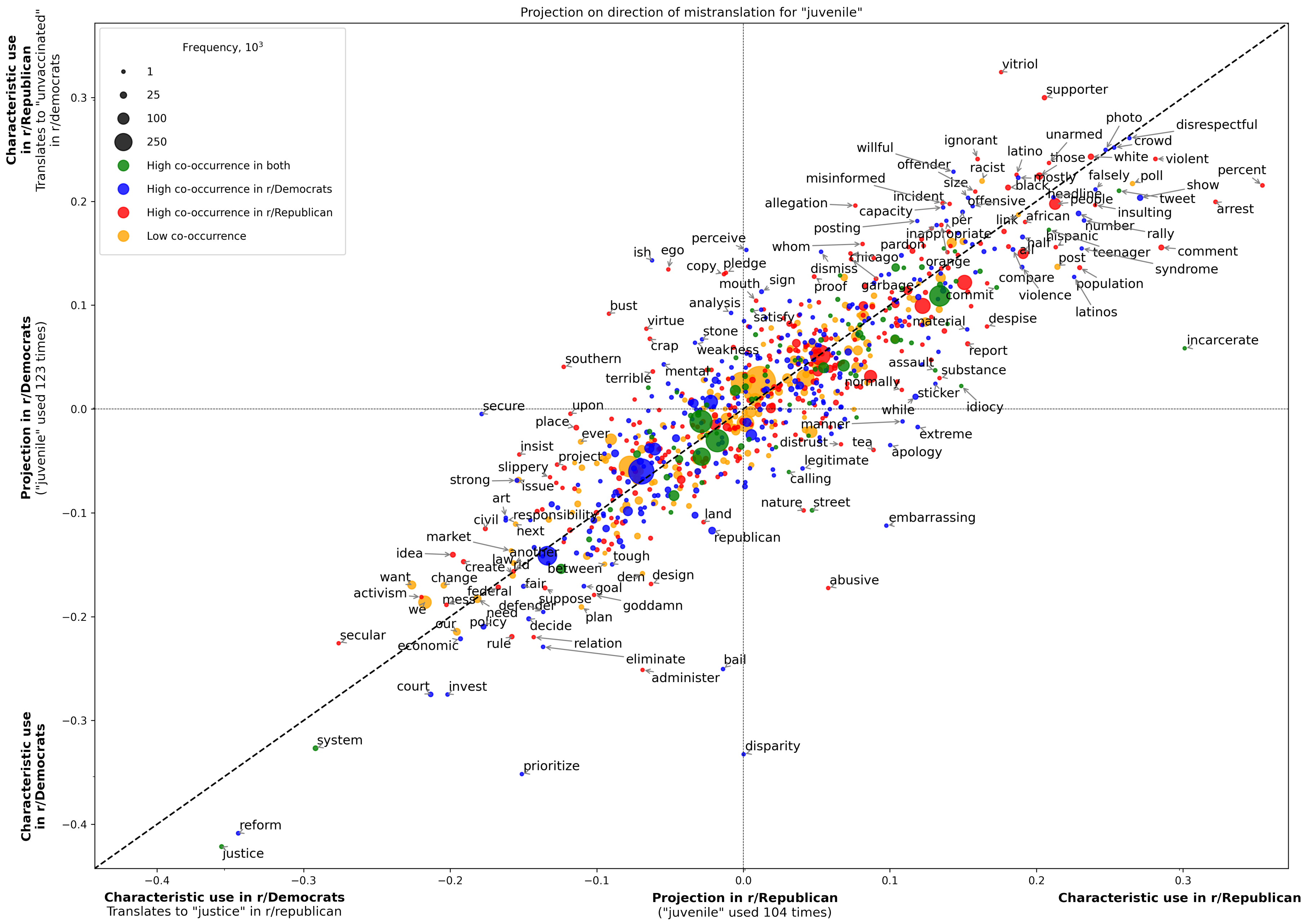

Figure 6: Projection of words on the offset for the embeddings of *juvenile*. The dispersion of words along the x-axis might appear greater than along the y-axis, as the figure has been stretched to the page. Words are coloured according to their co-occurrence with *juvenile*; see Eq. 1 for the definition of high co-occurrence.

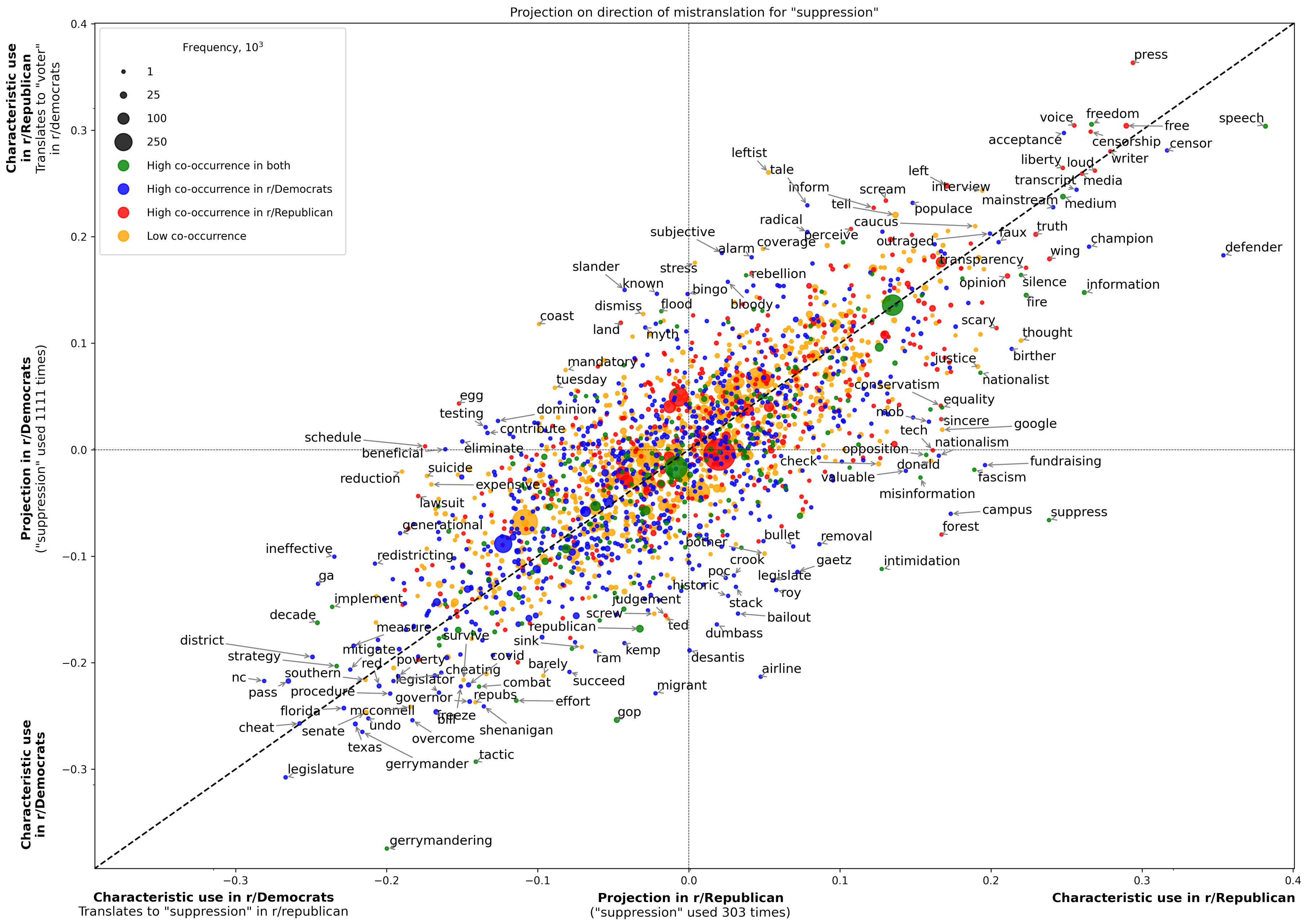

Figure 7: Projection of words on the offset for the embeddings of *suppression*. The dispersion of words along the x-axis might appear greater than along the y-axis, as the figure has been stretched to the page. Words are coloured according to their co-occurrence with *suppress*; see Eq. 1 for the definition of high co-occurrence.

| Most common words in the characteristic use of r/Republican | Most common words in the characteristic use of r/Democrats |
|---|---|
| company | supporter |
| illegal | comment |
| spending | court |
| marriage | bernie |
| subreddit | loan |
| reddit | judge |
| woman | r |
| kasich | day |
| police | democratic |
| biden | change |
| covid | hillary |
| study | end |
| immigration | sanders |
| corrupt | seat |
| government | his |
| immigrant | sander |
| cruz | stop |
| dollar | election |
| law | propaganda |
| budget | he |
| ask | majority |
| paul | volunteer |
| romney | lose |
| fetus | expand |
| vaccinate | voter |
| officer | scotus |
| city | gerrymander |
| serious | governor |
| disease | policy |
| freedom | white |

Table 2: Words that most unanimously project heavily towards the characteristic use in r/Republican (mean projection greater than 0.2) or in r/Democrats (mean projection less than -0.2).

Here, we have presented a new solution for how to overcome this methodological challenge. By focusing on the relative use of words within the corpora studied, we show how comparing projections along the direction of difference in the embedding space captures the most characteristic differences between the communities, no matter how minuscule this difference might quantitatively appear. At the same time, our approach also offers a novel way to measure difference in use, which complements the tendency for cosine distance and mistranslations to pick out low-frequent and polysemous words. By measuring the degree to which the characteristic uses in the dialectograms are also unique to each community, our measure of sense separation identifies words that are used most distinctly by the communities.

Our analysis illustrates how the combination of these methods can enable an unsupervised exploration of the discursive differences between two communities. First, measures of the degree to which words are embedded differently can suggest words that the two communities might use differently. Second, by constructing ideolectograms of these words, researchers can explore what the discursive differences consist of. While we have applied this approach to two contemporaneous political corpora, the same approach could also be used to compare other types of corpora or to explore how discourse evolves over time. Further, while we have compared two discursive communities, the approach could be extended to compare multiple communities, by aligning the multiple embeddings into a single, shared space.

While applying our methods to the case of US partisan subreddits reveals stark affective polarisation, we acknowledge a couple of important limitations. First, while we interpret dialectograms like Figure 5 to document derogatory out-group talk, what we observe is the saliency of such words in the context of out-group politicians. Technically, this could include cases of denying the application of the derogatory words to out-group members - although in that case, we would expect a word like 'not' to project in similar ways. Nonetheless, it is not completely evident to what extent computational maps of digital discourse, like the ones presented here, support the interpretation of the underlying meaning of the discourse (for a discussion, see Lee and Martin [2015] and the ensuing discussion in American Journal of Cultural Sociology vol 3, issue 3). In addition, as with all found or observational data, one should be careful when generalising the patterns identified [Salganik, 2018]. The active users of Reddit might systematically differ from supporters of the Republican and Democratic parties at large. Further, even if we are correct in assessing that the users of the subreddits belong to the more left-leaning fractions within each party (see the discussion around Figure 5), the users might still not be representative of the corresponding left-leaning voter segments. To probe the relations between observational and experimental data, further research could assess the degree to which survey-based approaches result in similar associations as those identified in the unsupervised examination of observational data (in the case of stance detection, see Joseph et al. [2021]).

In addition to these sampling and design issues, we encountered three technical issues that we did not manage to fully solve. First, our approach relies on the partition of text into distinct corpora. While many online platforms for communication provide partitions, such as the organisation of Reddit into subreddits, these do not necessarily correspond to homogeneous strands of discourse. We approached this by excluding users fairly active in both corpora, but one could also attempt to computationally partition the documents within a single corpus based on the distribution of words within it, e.g. using higher-order matrix factorisation of word-word-author matrices. Unlike topic modelling, which attempts to split documents into separate topics, this would rather partition the documents into two (or more) parts, that each speaks to the same set of topics, but in different ways.

Second, exploring the development of discursive alignment over time provides an obvious extension. We attempted to do so, by converting our static GloVe embedding to contextual representations of every word (token) in every time-stamped comment [Khodak et al., 2018, Rodriguez et al., 2022]. This approach, however, amplified the correlation between cosine distance and frequency; partitioning the comments into 19 periods, we found the cosine distance between the mean embedding of a word in each period correlates negatively with the mean frequency of the word in each period (median correlation of -0.75). Aggregating across all words, the median cosine distance between words showed an almost perfect negative correlation with the total number of words posted in the period (correlation of -0.98), suggesting the unlikely conclusion that the later periods with more activity saw less difference in use.

Third and finally, as discussed in the Supplementary Information, we found that converting contextual representations to static embedding provided quite rigid centroids, as measured by the perfect self-translation between our two corpora, even when we continue the masked language modelling. While we believe contextual representations provide a promising path for this type of analysis and could provide a way of studying variation over time, the implications of the pre-training corpus and architecture of language modelling are still unclear.

# 4 Materials

To illustrate the use of dialectograms, we collected text from supporters of the Democratic and the Republican party on the social media platform Reddit. Reddit is a platform structured around numerous topical message boards, so-called subreddits, where users post and comment on messages related to the subreddit's topic. On Reddit, support for the Democratic and the Republican party is centred around the two openly partisan subreddits r/Democrats and r/Republican [r/Republican, 2022, r/Democrats, 2022]. We picked these subreddits, as we expected them to contain English-language comments on similar topics, but based on different opinions and conceptions of politics.

We collect all comments made in the two subreddits from the Pushshift Reddit archive [Jason Micheal Baumgartner, 2022], using the PMAW API wrapper [Matthew Podolak, 2022]. The initial dataset contains 903.024 comments from r/Democrats and 1.038.151 comments from r/Republican, covering the period from January 2011 to September 2022. We drop 142.909 comments from r/Democrats and 106.239 comments from r/Republican, that were either marked as deleted by the user or removed by a moderator.

Having collected the comments, our aim is to create two corpora, based on which we can assess how the two different subreddits use the same words differently. While part of the differences between speech communities have to do with discourse, that is specific to one or the other, that is not our focus here. In fact, we will focus on the exact opposite; how the two communities put the same vocabulary to different uses. Hence, we take four overall steps to clean the comments and make the vocabularies identical: we remove repetitive comments; we process the comments to remove minor lexical variation; we restrict each corpus to users who are active in mainly one of the subreddits; and finally, we limit the vocabularies to words that appear at least a 100 times in each corpus. Below, we cover each in turn.

### Removing repetitive comments

First, our embedding-based approach is sensitive to the repetition of specific word associations, so we take steps to remove repetitive comments from our corpora. Overall, these comments can be split into two types. On one hand, Reddit has an eco-system of bots, that comment on the subreddits, using almost exactly the same language across different interactions. In many cases, these bots leave a signature in their comments, declaring themselves as such. On another hand, certain users may repeatedly post completely or near-identical comments; this is e.g. the case of moderators as well as some users posting information about how and where to vote during election campaigns (particularly evident on r/Democrats).

To identify such comments, we employed two approaches. We first read through comments containing certain keywords (starting with 'bot' and then branching out to other words and characters often appearing in the identified bot signatures, such as 'beep' and '^^'). In these comments, we identified highly distinct substrings (e.g. "*beep. boop.* I'm a bot") and removed all comments containing any of these substrings. In addition, we use co-occurrence information to identify words that exhibit unusual distributions. Specifically, we plot the frequency of words against the share of the vocabulary with which they co-occur. Words that co-occur with a relatively low share of the vocabulary compared to other words of similar frequency are targets for being repeated. By subsequently reading through comments including these words, we further identify repeated comments. With this procedure, we further remove 28.449 comments from r/Democrats and 69.231 comments from r/Republican.

### Comment preprocessing

Second, we then perform a range of preprocessing steps to 'clean' the data further and to make the vocabularies for each of the two corpora reasonably simplified and similar: we remove a few duplicate comments based on their id; convert emojis to text[8]; remove the parts of comments that quote other comments[9]; replace URLs with their domain[10]; remove usernames; align the use of quotation marks; expand contractions[11]; and convert hyphens, brackets, and parentheses to whitespace. Then we use spaCy to tokenize and lemmatise comments, as well as to obtain the part-of-speech for each token. Finally, we convert all tokens to lowercase, remove any left-over possessive "'s" suffixes from tokens, as well as tokens containing whitespaces or non-alphabetic characters, and remove comments with only one token, as no word associations can be learned from this.

While it is computationally expensive to assess how each of these preprocessing steps affects our results, we believe they all serve the purpose of removing minor lexical variation and ensuring that our corpora are maximally comparable at the level of the vocabulary, which is particularly important for the embedding models that do not rely on subword tokens, as they treat words as equally distinct regardless of their lexical similarity.

### Active and distinct users

Third, our inductive analysis relies crucially on the sorting of texts into two distinct corpora. The open nature of Reddit however implies that the scraped comments do not necessarily make up closed, homogeneous strands of republican and democratic discourse. In fact, against such an echo-chamber-based view, scholars are increasingly considering how user-based polarisation might be driven by the increased exposure to foreign opinions, that social media and forums like Reddit can generate [Törnberg, 2022]. Based on sampling and reading comments made by users who post in both subreddits, we indeed confirm that some degree of cross-partisan debate takes place. Regardless of its polarising effects, such cross-posting distorts our comparison to the extent that partisan users deploy the same partisan discourse in both subreddits, as it makes the use of words appear more similar across subreddits.

[8] Based on the emoji package.

[9] These quotes can be distinguished from ordinary quotes, as they are represented differently in the scraped text. We remove them as they too make comments repeat, in some cases with the reply being significantly shorter than the quoted text.

[10] Based on the tldeextract package.

[11] To do so, we rely on a range of custom-made regex-expressions.

To make the corpora more homogeneous we partition users based on their activity and generate the final corpora based on this user partition. Figure 8 shows how total user activity across the two subreddits varies with the share of the comments users post in one or the other. Based on this, remove comments from users, that have either posted less than 10 times or posted less than 90% of their comments in one of the two subreddits. This amounts to excluding 219.706 comments from r/Democrats and 285.802 from r/Republicans, which means that we conduct the analysis on a final preprocessed dataset of 511.906 comments from r/Democrats and 576.872 comments from r/Republican.

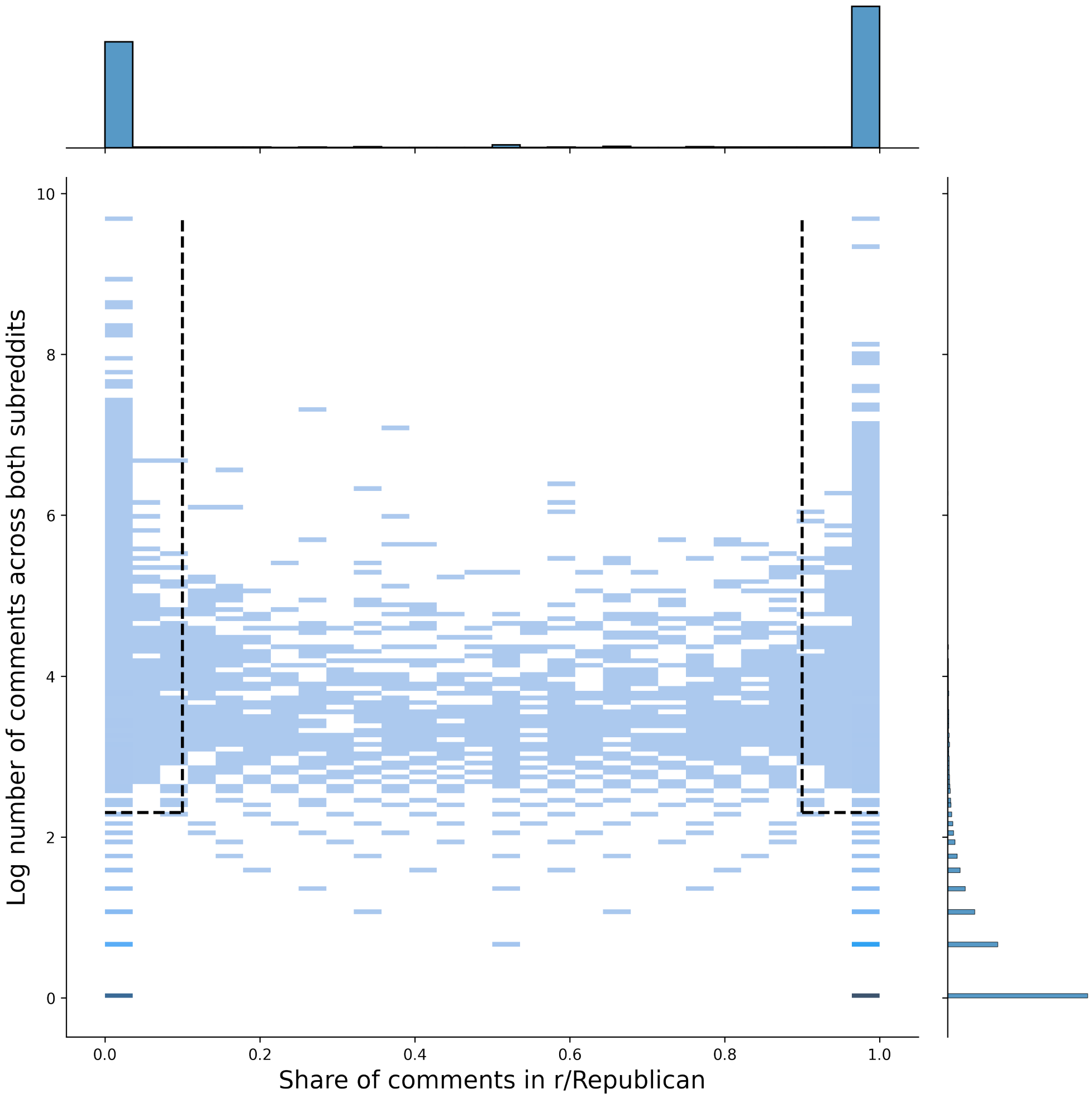

Figure 8: The log number of times a user posts in the two subreddits combined (y-axis) versus the share of these comments posted in r/Republican (x-axis). Most users post once, and (partially for that reason) most users post only in one of the two subreddits. There is however some degree of activity by users posting moderately and in both subreddits, indicating the kind of cross-commenting that can distort the comparison of the embedding spaces. The dashed lines indicate the users, whose comments make up the final dataset.

### Limit the vocabulary

The fourth and final step is to make two identical vocabularies based on the preprocessed datasets. To ensure identical vocabularies and that our interpretations of differences are based on a reasonable number of observations, we remove words from both corpora, that do not occur at least 100 times in each corpus. This leaves us with the same vocabulary of 5.738 unique words in each corpus.

# 5 Methods

Here, we describe the dialectograms and the sense separation measure in greater detail. In particular, we describe the GloVe embedding, including the hyperparameters we use to train them; the alignment methods we considered; the measures of difference in use, including the sense separation measure; and our attempts to remove the correlation between aligned cosine distance and word frequency.

### GloVe Embeddings

GloVe embeddings are based on minimising the objective

$$\sum_{i=1}^{N}\sum_{j=1}^{N} f(X_{ij}) \cdot [w_i \cdot \tilde{w}_j - [\log(X_{ij}) - b_i - \tilde{b}_j]]$$

where $X_{ij}$ is the co-occurrence count between word i and j; $w_i, \tilde{w}_j$ are the word and context vectors of word i and j, each of a pre-specified dimension D; $b_i, \tilde{b}_j$ are the word and context biases of word i and j; N the number of words in the vocabulary; and $f(\cdot)$ a weighting function, which by default is

$$f(x) = \begin{cases} (x/x_{max})^{\alpha} & \text{if } x < x_{max} \\ 1 & \text{otherwise} \end{cases}$$

for some specified values of $x_{max}$ and $\alpha$. The final embedding of word i is taken to be the sum of its word and context vector, $w_i + \tilde{w}_i$ [Pennington et al., 2014].

Intuitively vectors and biases are trained, such that the inner product between word and context vectors, $w_i \cdot \tilde{w}_j$, represent the degree to which the words co-occur beyond what would be expected based on their marginal frequencies, $\log(X_{ij}) - b_i - \tilde{b}_j$.[12] However, for word-pairs that co-occur less than $x_{max}$ times, the error in representation is down-weighted according to $(x/x_{max})^{\alpha} < 1$, so that in the limit of word-pairs that do not co-occur, $X_{ij} = 0$, no loss is obtained. As such, GloVe embeddings are by default not trained to represent non-occurrence.

We use the GloVe implementation made available by the authors of the model[13]. For all embeddings, we construct co-occurrence statistics based on a window size of 10 and train embeddings with 300 dimensions for 30 epochs. All other parameters are set to their default. We normalise the embedding of each word to have unit length, both before and after performing the alignment of the embeddings, equivalent to using cosine distance as opposed to euclidean distance for aligning and translating. We also apply the frequency adjustment described in the section on frequency correlation below, before we align the embeddings.

### Aligning Embeddings

The dialectograms rely on two aligned embedding spaces. To obtain aligned embedding spaces, several options exist; it is e.g possible to use the first embedding space to initialise the second. Here, we instead consider ways of aligning independently trained embeddings, $E_1$ and $E_2$, which we assume are both $N \times D$ matrices, sorted such that each row corresponds to the different embeddings of the same word.

Aligning word embeddings was already addressed in relation to the Word2Vec model, where the authors in a follow-up suggested learning a translation matrix, $W \in \mathbb{R}^{D \times D}$, from one embedding to the other [Mikolov et al., 2013b]. Specifically, they solve the following optimisation problem by gradient descent:

$$W_{GD} = \arg\min_{W} \| E_1 - E_2 W \|_F^2$$

where $\| \cdot \|_F^2$ is the squared Frobenius matrix norm.

The unrestricted nature of the optimisation implies, however, that $W_{GD}$ not only aligns $E_2$ to the space of $E_1$ but also distorts the internal relations between the embeddings of words in the $E_2$ space itself. To see this, consider the matrix of word-to-word inner products, given by

$$E_2 W_{GD} [E_2 W_{GD}]^T = E_2 W_{GD} W_{GD}^T E_2^T \neq E_2 E_2^T$$

[12] We expect the biases to capture the marginal word frequencies, which we confirmed is the case, as they exhibit perfect positive correlation.

[13] https://nlp.stanford.edu/projects/glove/

as long as $W_{GD}W_{GD}^T \neq I$, which is not guaranteed by the optimisation.

One way to impose this restriction is to require that W is orthogonal, i.e. that $WW^T = W^TW = I$. This amounts to requiring that W is a (potentially improper) rotation matrix. In this case, the optimisation problem is known as the Orthogonal Procrustes Problem, and referred to as Procrustes Analysis (PA). The problem was solved analytically in 1966 using Singular Value Decomposition (SVD) Schönemann [1966] (see also Artetxe et al. [2016]). In general, the SVD of a real-valued matrix X is the decomposition $X = USV^T$, where the left and right-singular matrices, U and V are orthogonal. Given this, the solution to the Orthogonal Procrustes Problem is:

$$\begin{aligned} E_2^T E_1 &= USV^T \\ W_{PA} &= \arg \min_{W, WW^T = I} \parallel E_1 - E_2 W \parallel_F^2 \\ &= UV^T \end{aligned}$$

where the first line is the SVD of the matrix of inner products between the dimensions of the two embedding spaces. In this case,

$$E_2 W_{PA} [E_2 W_{PA}]^T = E_2 W_{PA} W_{PA}^T E_2^T = E_2 U V^T V U^T E_2^T = E_2 E_2^T$$

due to the orthogonality of U and V. The constrained nature of the optimisation implies that $W_{PA}$ provides a worse fit than $W_{GD}$, but it keeps the structure of the transformed embedding intact.

In performing the alignment, PA will put greater emphasis on aligning dimensions with greater variation or length, as measured by the L2 norm of each dimension j, $\parallel E_{\cdot,j} \parallel_2$. For this reason, it is generally common to demean and unit-normalise the variation of each column before proceeding. As we observe our embeddings to have fairly similar lengths, this is unlikely to play a major role, but it does motivate a third suggestion for how to align embeddings spaces, Canonical Correlation Analysis (CCA).

Whereas PA aligns dimensions based on their scale-dependent inner products, CCA aligns dimensions based on their correlation. As such, CCA can be performed using SVD on the correlation matrix (instead of the inner product matrix, $E_2E_1^T$) between dimensions. However, calculating the correlation matrix is not necessary - it is instead possible to perform CCA as a two-step procedure based on $E_1$ and $E_2$ directly, which in turn clarifies the relation between PA and CCA William [2011].

Unlike the PA procedure presented above, CCA is typically done by transforming both embeddings, that is, we are looking for two transformation matrices, $W_{CCA_1}$ and $W_{CCA_2}$, transforming $E_1$ and $E_2$ respectively. First, note that based on the SVD of each embedding $E_k = U_k S_k V_k^T$, we can rewrite the SVD of $E_2^T E_1$ used in PA as:

$$E_2^T E_1 = V_2 S_2^T U_2^T U_1 S_1 V_1^T$$

While PA aligns embeddings based on the inner product of the embedding dimensions, $E_2^T E_1$, CCA aligns embeddings based on the inner products of their left singular vectors, $U_2^T U_1$. The left singular vectors are orthogonal matrices, i.e. uncorrelated and scaled to unit length. As such, CCA can be interpreted as aligning the unweighted principal components of the two embedding spaces, with transformation matrices given as:

$$\begin{aligned} U_2^T U_1 &= U_* S_* V_*^T \\ W_{CCA_1} &= V_1 S_1^{-1} V_* \\ W_{CCA_2} &= V_2 S_2^{-1} U_* \end{aligned}$$

The simultaneous transformation of both embeddings spaces has the additional benefit that the dimensions of the transformed spaces have the interpretation of being sorted according to their correlation - the embeddings have the highest correlation along the first dimension, the second highest along the second dimension and so forth. This is, however, not a feature unique to CCA, as it is possible to reformulate PA to have a similar interpretation, in which the two transformation matrices are

$$\begin{aligned} E_2^T E_1 &= USV^T \\ W_{PA_1} &= V \\ W_{PA_2} &= U \end{aligned}$$

In this version of PA, the transformed dimensions are not sorted by their correlation, but by their inner product. Inspecting this further, V has the interpretation of projecting from the $E_1$ space to the shared space sorted by inner product, while $V^{-1} = V^T$ projects the opposite way (likewise for U and $E_2$). Comparing this simultaneous version of PA to the single $W_{PA} = UV^T$ shows that $W_{PA}$ performs two projections, first projecting $E_2$ to the sorted inner product space by U, after which $E_2U$ is further projected to the $E_1$ space by $V^T$. This twofold version of PA alignment is what we use here.

In some applications, the embeddings are aligned based on a subset of words, such as high-frequent stop words, that are assumed to be used similarly across the corpora. In our case, we align the embeddings based on all words, to avoid making assumptions about the similarity prior to the analysis.

### Identifying words used most differently

In the empirical application, we consider three measures of difference in use. The most common measure of embedding distance is cosine distance (CD), which we also find to work well in our validation task:

$$CD(w_{1,i}, w_{2,i}) = 1 - \frac{w_{1,i} \cdot w_{2,i}}{\| w_{1,i} \|_2 \| w_{2,i} \|_2}$$

In addition to cosine distance, prior research on inductively discovering differences between contemporary speech communities has focused on mistranslations [KhudaBukhsh et al., 2021, Milbauer et al., 2021]. A word mistranslate from $C_1$ to $C_2$ if the nearest neighbour of $w_{1,i}$ in the aligned embedding of $C_2$ is not $w_{2,i}$, and similarly from $C_2$ to $C_1$; here, we consider a word to mistranslate, if either of the directions does not translate to the word itself.

Sorting mistranslating words by descending frequency partially counters the tendency for cosine distance to identify low-frequent words. Mistranslation is, however, sensitive to highly co-occurring words, such as in the case of first and last names (in our application, *warren* e.g. mistranslates to *elizabeth*). One remedy could be to instead consider a word as mistranslating if it is not among its own k-nearest-neighbours for some value of $k > 1$. Yet, in this case, few or no words will likely mistranslate at all; word embeddings attempt to compress the full variety of discourse into a dense space, and against this background, words are overall likely to be used much more similarly than differently.

Further, by inspecting the dialectograms for words, that exhibit high cosine distance or are among the highest-frequent mistranslations, we observe several cases of polysemy; examples include echo *chamber* versus *chamber* of commerce, the *bell* company versus an alarm, the *viral* load versus a *viral* video, and *turkey* as a country versus an animal. We also observe cases where two persons share part of a name, such as Joy and Harry *reid*. In these cases, both subreddits are likely to use such polysemous words in both senses, but to varying degrees. This observation is what motivates our sense separation measure, which attempts to identify words, where the characteristics use of each corpus is also relatively unique to it.

In particular, for any focal word i, we identify the words in each corpus, that co-occur more with the focal word than the product of their marginal frequencies would suggest, i.e. more than we would expect if co-occurrence was independent. If $C^k_{i,j}$ is the co-occurrence count between focal word i and context word j in a corpus k, $N^k_c$ is the sum of all co-occurrence counts in that corpus and $N_w$ is the number of words, the criteria is:[14]

$$EC^k_{i,j} = \frac{C^k_{i,j} \cdot N^k_c}{\sum_{h=1}^{N_w} C^k_{i,h} \cdot \sum_{h=1}^{N_w} C^k_{h,j}} > 1 \tag{1}$$

Given the sets of words co-occurring highly with the focal word in each corpus, we identify the subset of these words, that only co-occur highly in one of the corpora, but not both, $HC^1_i = \{j \in \{1,..,N_w\} \mid EC^1_{i,j} > 1 \wedge EC^2_{i,j} \leq 1\}$ for the first corpus and likewise for the second. Intuitively, these words mark the associations with the focal word that are distinct for each corpus.

For each set of words, $HC^1_i$ and $HC^2_i$, we calculate the mean of their scalar projections ($\alpha^1_{i,j}$ and $\alpha^2_{i,j}$) on the offset for the focal word - in the case of $HC^1_i$,

$$\overline{HC^1_i} = \frac{1}{|HC^1_i|} \sum_{j \in HC^1_i} \frac{\alpha^1_{i,j} + \alpha^2_{i,j}}{2} \in [-1, 1]$$

and likewise for $HC^2_i$. Taking the average of the scalar projections is equivalent to projecting the words onto the dashed line running along the main diagonal in the dialectogram; taking the mean of these averages hence captures the overall position of the uniquely high co-occurring words along the main diagonal. Finally, to get our measure of sense separation we subtract the two means,

$$S_i = \overline{HC_i^1} - \overline{HC_i^2}$$

[14] This is equivalent to cases where the pointwise mutual information (PMI) between the words is positive.

### Alignment and frequency

As Figure 3a shows, CD is negatively correlated with frequency. Controlled experiments suggest that this is a feature of the embedding models; they find a large share of the correlation between semantic change and frequency to reappear in settings, where none would be expected [Dubossarsky et al., 2017]. The authors argue that this is a general feature of count-based models, as the expected cosine distance between two samples drawn from the same multinomial distribution (representing a given co-occurrence distribution) decreases with the size of the samples.

Experimenting with the frequency weight of the GloVe objective function, we also conclude that the correlation is an artefact of the models. In particular, by either removing the weighting function, (equivalent to weighting all pairs equally) or by inversing the weighting so that lower-frequent pairs receive a higher weight than high-frequent pairs, we observe that the correlation disappears. This, however, come at the cost of drastically increasing the share of mistranslations (from 11% to 67%), why we stick to the default weighting scheme.

In addition to adjusting the Glove model, we also attempted to remove frequency information from the subsequently obtained embedding spaces, before they are aligned. Here, methods have been developed to identify a subspace, that correlates with a semantic feature of interest [Rothe et al., 2016, Rothe and Schütze, 2016, Dufter and Schütze, 2019]. After identifying such a subspace, the embedding can be projected to the complement of the subspace, to 'remove' variation corresponding to the corresponding semantic feature.

To obtain a subspace that should capture variation in frequency, we calculate the inner product between the log frequency of words in corpus k, $f_k \in \mathbb{R}^N$ and the dimensions of the embedding space,

$$w_k = E_k^T f_k \in \mathbb{R}^D$$

Intuitively, this (row) vector should point in the direction that overall varies most with frequency. To remove this direction from the embedding space, we project the embedding to the orthogonal complement of this direction,

$$E_k^* = E_k[I - w_k^T w_k]$$

where $E_k^*$ is the adjusted embedding.

# 6 Model and Measure Selection

Over the last 10 years, the NLP community has produced a long range of neural network architectures for learning word embeddings. At a high level, an important distinction is between static and contextual embedding models; whereas static embeddings provide a single vector for each word (type), contextual embeddings provide a numerical representation for each time a word (token) appears. Overall, contextual embeddings provide richer representations, allowing for comparison of how a word is differently distributed as opposed to only differently positioned. However, the contextual embedding models are typically more complex and trained on corpora larger than many of social scientific interest.

For this reason, it is advised to use a contextual model pre-trained on a larger corpus, instead of training one the corpus of interest from scratch [Rodriguez and Spirling, 2022]. Although a pre-trained model can be calibrated to a given corpus by continuing the language modelling, it still creates a trade-off; either using the simpler representations and measures of alignment from the corpus-specific static embeddings or using the richer representations and measures from the only partly-attuned contextual embedding [Manjavacas and Fonteyn, 2022].

In addition, several ways of measuring differences between the resulting embeddings also exist. This raises the question of a) what embedding models and measure works best and b) whether such an optimal approach is reliable enough to serve as a computational base for the interpretation of the corpora. This is a particularly pressing challenge for computational inductive approaches, as they cannot easily rely on the train-validate-test partition behind many supervised approaches.

Computational models and measures are typically evaluated by assessing how well they perform in a controlled setting, which establishes how 'correct' or 'true' performance looks. In the case of word embedding models, such evaluation can at a high level be divided into extrinsic and intrinsic approaches. Extrinsic (or downstream) evaluation consists in using the embedding model to perform a different, but related task (typically a classification task such as part-of-speech tagging or named entity recognition). This gives an indirect assessment of the quality of the word representations. Alternatively, it is also possible to devise an intrinsic task, in which the properties of the embedding space are more directly probed. Given our corpora-focused approach, in which our interpretations rely crucially on the properties of the embedding model and measures of difference, we opt for an intrinsic evaluation.

Most embedding models are based on optimising a loss function, which serves as an intrinsic measure of the quality of the embedding. This is however not always the case (such as the SVD embeddings) and loss levels are often not comparable across models, why embedding models are often evaluated intrinsically by other means. One intrinsic option is to compare a measure of how confident the embeddings are in predicting the co-occurrences of the training data, such as perplexity. Another intrinsic option is to compare the associations of the embedding model to equivalent associations made by humans, such as in word intrusion tasks or from surveys [Chang et al., 2009, Murphy, Brian et al., 2012, Kozlowski et al., 2019]. A third intrinsic option, akin to an extrinsic evaluation, is to evaluate embedding models by how well they perform on word analogy tasks, such as the Google Analogy Test Set or Bigger Analogy Test Set (BATS) Mikolov et al. [2013a], Gladkova et al. [2016].

The last two approaches conceptually mix up a corpora-focused evaluation - how well does the embedding model represent the associations of the corpora at hand - with a model-focused evaluation - how well does the model represent general or objective associations, not tied to a particular corpus. To the extent the two evaluation objectives correlate, such as both improving when working with larger corpora, the model-focused evaluation provides an efficient evaluation method, but it does not constitute an ideal evaluation from a corpora-focused perspective.

In line with the basic intuition of creating a validation test, that resembles the purported use of the models and measure, while establishing a basis for adjudicating how well the models and measures perform, we create a synthetical corpus, that resembles an actual corpus in most regards, except that it has certain words swapped. By controlling which words are swapped, as well as the degree to which they are swapped, we can evaluate to what extent the embeddings and measures correctly identify the degree to which a given word is swapped, as well as which word it was swapped with.

Specifically, we create a partially swapped copy of a smaller version of the r/Republican corpus, in which 600 words are swapped (∼11% of vocabulary).[15] To ensure words are swapped in a somewhat syntactically sound manner while allowing us to evaluate how our models and measures perform at different word frequency levels, we swap words based on the following procedure:

- Words are partitioned into 10 frequency deciles.
- For each decile, we randomly draw 30 pairs of words without replacement, such that the two words making up each pair belong to the same part-of-speech.
- To each word pair we assign one of ten degrees to which they will be swapped (10%, 20%, ..., 90%, 100%), such that for each frequency decile and swap degree, we obtain three word pairs, where the words in each pair belong to the same part-of-speech.
- Based on this, we create a swapped version of the small r/Republican corpora, in which each word in each pair is swapped with its paired word according to the allotted degree.[16]

Given the original and swapped r/Republican corpora, we apply three word embedding models: the skip-gram with negative sampling (SGNS) version of Word2Vec; Global Vectors for Word Representation (GloVe); and a distilled version of a pre-trained RoBERTa model [Mikolov et al., 2013a, Pennington et al., 2014, Liu et al., 2019, Sanh et al., 2020]. In addition to separate DistilBERT embeddings, we also train a joint DistilBERT model, by continuing the masked language modelling on both corpora at once, inserting a corpus-specific token at the beginning of each comment.

We evaluate how well each of a range of applicable measures described below performs in sorting words according to their degree of swap, by measuring the Spearman's (rank) correlation between the distance measures and the swap degrees.[17,18]

[15] The smaller dataset was collected at the end of 2020. We later updated the dataset to have a wider coverage. The smaller r/Republican corpus is roughly half the size of the final.

[16] While both words in a pair belong to the same frequency decile, their frequency still differs. Instead of opting for a fixed number of swaps for each pair, in which case we could only attain the allotted swap degree on average, we swap each word in the pair to the allotted degree, with the implication that frequencies of the swapped words change.

[17] We use the spearmanr function from the scipy.stats module

[18] In addition to the swap degree, the cosine distance between the swapped words (from the original embedding) is likely to affect the expected distance - the further apart two words are, the more should swapping them induce distance between their aligned

Before we proceed, we note that this validation task is not perfect. Most importantly, real-world corpora will likely be much more different than the corpora considered here, where most word associations are the same. Further, while we have not done any hyperparameter tuning based on the validation task, this is partly because it is likely not suitable for certain hyperparameters. In the case of window size, for instance, smaller windows will likely make it easier to identify differences in this simplified setting, without necessarily generalising to our use case. All in all, we still believe it provides a useful indication of the relative performance of the various approaches.

## Training embeddings

For training the SGNS embeddings, we use the implementation made available by Gensim: https://radimrehurek.com/gensim/models/word2vec.html. For training GloVe embeddings, we use the implementation made available by the authors of the model: https://nlp.stanford.edu/projects/glove/. In both cases, we train 300-dimensional embeddings for 30 epochs with a window size of 10. All other parameters are set to the default of the implementations.

For DistilRoBERTa, we rely on the pretrained distilroberta-base checkpoint made available on Huggingface: https://huggingface.co/distilroberta-base, as well as the various modules for training models (AutoModelForMaskedLM, DataCollatorForLanguageModeling, Trainer, TrainingArguments). To extract embeddings for words in comments, that are longer than the token limit of 512, we split up the tokenized comment into separate parts, such that each part except the last has 512 tokens, but with an overlap of 50 tokens in each end of the parts; the 50 last tokens of the first part are identical to the first 50 tokens of the second part, the last 50 tokens of the second part are identical to the first 50 tokens of the third part and so on. For tokens that end up in the overlaps, and hence appear twice, we take the average of the embeddings from each part. For the joint model, we add the corpus-specific tokens to the AutoTokenizer of the pretrained checkpoint.

## Measures of difference

Given two (aligned) embedding spaces, we consider several approaches to measure their differences. At a high level, we distinguish between methods applicable to static representations (including the static representation created based on contextual embeddings) and methods applicable to contextual embeddings.

### Static measures

For the static representations (including contextual centroids), we consider five measures of difference. In addition to cosine distance and our suggested measure of sense separation described above, these include:

**Nearest neighbour overlap:** It is possible to measure similarity without aligning the embedding spaces, by comparing the local structure of the embeddings. Here, we measure the similarities of the words neighbourhood in the two embeddings by comparing their overlap:

$$\mathrm{d}_{\mathrm{w}}^{\mathrm{k-NN}} = 1 - \frac{|\mathrm{v}_{\mathrm{C1,w}}^{\mathrm{k-NN}} \cap \mathrm{v}_{\mathrm{C2,w}}^{\mathrm{k-NN}}|}{\mathrm{k}}$$

where $\mathrm{v}_{\mathrm{C1,w}}^{\mathrm{k-NN}}$ is the set of k nearest neighbours for w in $\mathrm{C}_1$, and $|\mathrm{v}_{\mathrm{C1,w}}^{\mathrm{k-NN}} \cap \mathrm{v}_{\mathrm{C2,w}}^{\mathrm{k-NN}}|$ it the size of the intersection of the two neighbourhoods.

**PCA of word offsets:** In addition to the distance between the embeddings of a word, such as captured by $\mathrm{d}_{\mathrm{cos}}$, there could also be information in the direction of the difference between the embeddings. In the naive case of a single, repeated discursive difference, we should ideally be able to identify a single direction corresponding to this difference. While this is an improbable assumption, it motivates the idea of identifying the direction, that represents most of the differences between the two embeddings. One way to do so is to perform Principal Component Analysis (PCA) of the vector offsets, $\mathrm{O} = \mathrm{E}_1 - \mathrm{E}_2 \in \mathbb{R}^{\mathrm{N} \times \mathrm{D}}$ and use the absolute value of words' principal score on the first principal component as a measure of their difference:

$$\mathrm{O} = \mathrm{USV}^{\mathrm{T}}$$
$$\mathrm{d}_{\mathrm{w_i}}^{\mathrm{PCA}} = |\ \mathrm{US}_{[\mathrm{i},1]}\ |$$

representations in the original and swapped corpora. We hence also correlated the measures and models with the product of swap degree and initial cosine distance, which gave similar results.

where $US_{[i,1]}$ denote the ith row and 1st column of the principal scores, US and $|\cdot|$ is the absolute value.

**Distance to SVM boundary:** Another way to identify any shared differences between the embeddings is to train a Support Vector Machine (SVM) to separate them. SVM identifies the hyperplane that best separates the two embeddings (maximum margin hyperplane). Intuitively, the easier it is two separate a given word, the more likely is it used differently. As a measure of separability, we rely on the (euclidean) distance between the embedding of a word and the SVM hyperplane, which is equivalent to the length of the rejection vector between the embedding and the hyperplane, $r^{SVM}_{C_j,w}$ - for a given word, we obtain two such measures, corresponding to each of its two embeddings, which we sum:

$$d^{SVM}_w = \| r^{SVM}_{C_1,w} \|_2 + \| r^{SVM}_{C_2,w} \|_2$$

### Contextual measures

Extracting static embeddings from contextual embeddings amounts to collapsing the multiple contextual representations of a word across different contexts to a single point in the discursive space. Instead, we might consider comparing the full distribution of a word's contextual representations between the two corpora. Since the shared embedding space lends itself so easily to measures of distance, the most obvious candidate for comparing the difference in distribution is the Wasserstein distance (WD), which in the case of optimal transport is also known as Earth Mover Distance.

Earth Mover Distance suggests the perhaps most intuitive understanding of this measure; by considering the two probability distributions as two piles of earth, the distance between the distributions is then defined as the cost of rearranging the earth from one pile so that it is identical to the other pile, with the cost of moving earth from one point to another point defined as the distance between the points. The optimal transportation of earth (or probability mass) amounts to incurring the lowest cost.

In the case of text analysis, this measure was introduced as Word Mover Distance (WMD), aimed at measuring the similarity of documents [Kusner et al., 2015]. By considering documents as a distribution of its words in the embeddings space, the similarity of two documents can be measured as the least cost of moving the points from one document to the other. Subsequently, WMD was also used to measure the extent to which a document engages with one or more focal concepts (Concept Mover Distance), by measuring the cost of moving the embedded words of a document to the embedding of the focal concepts[Stoltz and Taylor, 2019].

In our case, we can use WD in a slightly different way, where we treat a word as two distributions, based on its contextual representations in each embedding, and calculate the WD between these two distributions, based on the cosine distance between the representations. Formally, if $N^{C_1}_w$ and $N^{C_2}_w$ denote the number of contextual representations of w in $C_1$ and $C_2$ respectively, the WD distance is defined as:

$$CD\left(v_{w_i}, v_{w_j}\right) = 1 - \frac{\langle v_{w_i}, v_{w_j}\rangle_2}{\| v_{w_i} \|_2 \| v_{w_j} \|_2}$$

$$T^* = \arg \min_{T \in \mathbb{R}^{N^{C_1}_w} \times \mathbb{R}^{N^{C_2}_w}} \sum_{i=1}^{N^{C_1}_w} \sum_{j=1}^{N^{C_2}_w} T_{i,j} \cdot CD\left(v^i_{C_1,w}, v^j_{C_2,w}\right)$$

$$\text{s.t} \sum_{i=1}^{N^{C_1}_w} T_{i,j} = \frac{1}{N^{C_2}_w}, \sum_{j=1}^{N^{C_2}_w} T_{i,j} = \frac{1}{N^{C_1}_w}$$

$$d^{WD}_w = \sum_{i=1}^{N^{C_1}_w} \sum_{j=1}^{N^{C_2}_w} T^*_{i,j} \cdot CD\left(v^i_{C_1,w}, v^j_{C_2,w}\right)$$

Experimenting with the WD measure, we noticed that increasing the size of the distributions to be matched leads to lower WD. To avoid a direct correlation between frequency and WD, we implement a sampling procedure. Specifically, for each word, we draw 20 samples without replacement of 75 contextual representations from each corpus. For each sample, we calculate the WD, and take the mean across the 20 WD estimates.

We implement WMD using the POT Python package.

### Joint contextual measures

Intuitively, for words used similarly, the corpus-tokens should carry no information, whereas, for words used differently, the difference in the embeddings of the corpus-tokens should help distinguish between the different uses of the word. Hence, similar in spirit to the approach taken in $d_w^{PCA}$ and $d_w^{SVD}$, we then create a measure based on the difference between the corpus-specific tokens in the following way. First, we obtain the contextual representations for each corpus-specific token, $\{v_i^{C_1}\}_{i=1}^{N_{C_1}}$ and $\{v_j^{C_2}\}_{j=1}^{N_{C_2}}$. Then we calculate the mean embedding for each set of corpus-tokens, $\bar{v}^{C_1} = \frac{1}{N_{C_1}} \sum_{i=1}^{N_{C_1}} v_i^{C_1}$ and likewise for $C_2$. For each word in the vocabulary, we then calculate the two centroids of its contextual representations in the two corpora, $\bar{v}_{C_1,w}$ and $\bar{v}_{C_2,w}$, and project these centroids onto the offset between the mean corpus-token embeddings, $\bar{v}^{C_1} - \bar{v}^{C_2}$, to obtain the normalised scalar projections equal to the cosine similarity (CS) between word centroid and corpus-token-offset

$$CS\left(\bar{v}_{C_1,w}, \bar{v}^{C_1} - \bar{v}^{C_2}\right) = \frac{\langle \bar{v}_{C_1,w}, \bar{v}^{C_1} - \bar{v}^{C_2} \rangle_2}{\| \bar{v}_{C_1,w} \|_2 \| \bar{v}^{C_1} - \bar{v}^{C_2} \|_2}$$

Finally, as our measure of difference, we then calculate the absolute value of the difference in the projections of the centroids (DPC),

$$d_w^{DPC} = |CS\left(\bar{v}_{C_1,w}, \bar{v}^{C_1} - \bar{v}^{C_2}\right) - CS\left(\bar{v}_{C_2,w}, \bar{v}^{C_1} - \bar{v}^{C_2}\right)|$$

## Validation results

Table 3 reports the correlations based on all words. In general, cosine distances and our suggested co-occurrence separation measure display the highest Spearman correlation with the swap degrees. Cosine distance based on the centroids from contextual embeddings does not correlate better than their static counterparts, but they do correlate much better when using the SVM distance and offset PCA measures, in which cases the static embeddings are useless. The nearest-neighbour measure works for static and dynamic centroids alike but is always correlating less than the corresponding cosine distance measure.[19]

The non-statics measures also correlate subpar compared to the static cosine measure. Cosine WMD correlate less than both cosine distance, offset PCA and 30-NN, while the projection difference from the joint embedding is more promising, despite still correlating less than the cosine distance.

| Embedding | Cosine distance | SVM distance | Offset PCA | 30 NN difference | Sense separation | Cosine WD | Projection difference |
|---|---|---|---|---|---|---|---|
| W2V, PA | 52.0 | 8.0 | 0.0 | 43.0 | 54.0 | – | – |
| GloVe, PA | 54.0 | 3.0 | 0.0 | 50.0 | 53.0 | – | – |
| W2V, CCA | 52.0 | 2.0 | 0.0 | 39.0 | 52.0 | – | – |
| GloVe, CCA | 50.0 | 3.0 | 0.0 | 37.0 | 46.0 | – | – |
| RoBERTa, pretrained, 4 last layer | 53.4 | 15.0 | 46.1 | 36.2 | 23.8 | 29.9 | – |
| RoBERTa, 5 epoch, 4 last layer | 52.1 | 29.6 | 45.6 | 40.6 | 23.0 | 30.2 | – |
| RoBERTa, 20 epoch, 4 last layer | 52.7 | 1.0 | 41.2 | 43.2 | 33.7 | 33.6 | – |
| RoBERTa, pretrained, last layer | 53.3 | 17.1 | 45.8 | 37.1 | 24.1 | 32.1 | – |
| RoBERTa, 20 epoch, last layer | 49.5 | 9.1 | 7.4 | 41.1 | 19.5 | 32.4 | – |
| Joint RoBERTa, 5 epoch, 4 last layer | – | – | – | – | – | – | 43.8 |
| Joint RoBERTa, 20 epoch, 4 last layer | – | – | – | – | – | – | 44.7 |
| Joint RoBERTa, 20 epoch, last layer | – | – | – | – | – | – | 44.8 |

Table 3: **All words:** Spearman correlation between the degree of swap and various measures (columns) for different embedding approaches (rows) and all words. Not all measures are applicable to all embeddings (marked with a dash).

[19] We did experiment with varying the number of nearest neighbours. There was not a single number that worked best for all embeddings, but even when choosing the optimal number of neighbours specific to each embedding, this measure still does not achieve higher correlation than cosine distance.

As the swap degree is zero for ∼89% of words in the vocabulary, the difference in correlation might to a high degree be driven by the sorting of non-swapped words. While correlation based on all words most closely mimics the explorative situation, we also perform the correlations for the swapped words only, to assess how well the measures work for words in which we expect a difference. Table 4 reports the correlation for swapped words only. Generally, correlations are higher and cosine distances achieve the highest correlation, with static embeddings this time correlating higher that the centroids from contextual embeddings. The sense separation measure does, however, not perform as well in this setting, suggesting that its strength is less in deciding the degree of swap, but rather whether words have been swapped or not.

| Embedding | Cosine distance | SVM distance | Offset PCA | 30 NN difference | Sense separation | Cosine WMD | Projection difference |
|---|---|---|---|---|---|---|---|
| W2V, PA | 92.0 | 6.0 | 2.0 | 67.0 | 56.0 | – | – |
| GloVe, PA | 91.0 | 1.0 | 0.0 | 83.0 | 47.0 | – | – |
| W2V, CCA | 92.0 | 2.0 | 4.0 | 66.0 | 49.0 | – | – |
| GloVe, CCA | 91.0 | 2.0 | 0.0 | 66.0 | 58.0 | – | – |
| RoBERTa, pretrained, 4 last layer | 77.7 | 53.0 | 46.8 | 46.4 | 24.2 | 51.1 | – |
| RoBERTa, 5 epoch, 4 last layer | 82.1 | 53.3 | 52.3 | 65.1 | 32.5 | 52.7 | – |
| RoBERTa, 20 epoch, 4 last layer | 82.5 | 40.0 | 33.2 | 72.4 | 38.8 | 60.2 | – |
| RoBERTa, pretrained, last layer | 75.3 | 52.6 | 44.2 | 51.2 | 21.2 | 51.8 | – |
| RoBERTa, 20 epoch, last layer | 82.0 | 6.8 | 32.8 | 66.4 | 34.7 | 58.4 | – |
| Joint RoBERTa, 5 epoch, 4 last layer | – | – | – | – | – | – | 46.7 |
| Joint RoBERTa, 20 epoch, 4 last layer | – | – | – | – | – | – | 35.2 |
| Joint RoBERTa, 20 epoch, last layer | – | – | – | – | – | – | 36.5 |

Table 4: **Swapped words only:** Spearman correlation between the degree of swap and various measures (columns) for different embedding approaches (rows) and swapped words only. Not all measures are applicable to all embeddings (marked with a dash). Embeddings are still aligned based on all words, and the nearest neighbours are identified among all words.

In addition to assessing how well the models and measures recover the induced differences, we can also ask how well the static embeddings identify which words a given word has been swapped with (if any). For words that are not swapped, as well as for words that are swapped less than 50%, we expect the word to translate consistently to itself (i.e. be its own nearest neighbour). For words that are swapped more than 50%, we expect words to translate to their paired word, while for words that are swapped exactly 50%, we expect the translation to be indecisive, that is in some cases translating to the word itself, while in others translating to its paired word.

Table 5 shows the result of this translation task. Here, there is a clear difference between the static embeddings and contextual centroids - while all embeddings correctly translate words that have not been swapped, as well as (most) words swapped less than 50%, only the static embeddings correctly translate (most) words swapped more than 50%. This is the case because the contextual centroids almost always translate consistently, regardless of the post-training approach.

| | Unswapped words | Less than 50% swap | More than 50% swap |
|---|---:|---:|---:|
| W2V, PA | 100.0 | 94.6 | 94.3 |
| GloVe, PA | 100.0 | 92.1 | 93.7 |
| W2V, CCA | 100.0 | 95.0 | 93.0 |
| GloVe, CCA | 99.6 | 82.5 | 80.0 |
| RoBERTa, pretrained, 4 last layer | 100.0 | 100.0 | 0.0 |
| RoBERTa, 5 epoch, 4 last layer | 100.0 | 100.0 | 0.0 |
| RoBERTa, 20 epoch, 4 last layer | 100.0 | 99.2 | 0.7 |
| RoBERTa, pretrained, last layer | 100.0 | 100.0 | 0.0 |
| RoBERTa, 20 epoch, last layer | 100.0 | 100.0 | 0.0 |

Table 5: Translation accuracy for different embedding approaches (rows) by the degree of swap (columns).

# Supplementary Information

## Top 100 words used most differently

The following tables show the top 100 words identified by each of the three measures of difference used here. The corresponding dialectograms for these words below can be found at https://github.com/A-Lohse/Dialectograms.

### Highest aligned cosine distance

| 0-20 | 20-40 | 40-60 | 60-80 | 80-100 |
|---|---|---|---|---|
| voluntary | 🤡 | homelessness | goldwater | massacre |
| juvenile | 🇺🇸 | gaslighte | ps | aide |
| com | url | org | additionally | divorce |
| 🤪 | lastly | helicopter | ffs | graph |
| gingrich | forbes_url | disclose | disrupt | cloud |
| sheet | 😡 | viral | facilitate | lewis |
| 👏 | me | object | gym | pee |
| settlement | broadband | pipeline | mccarthy | djt |
| keystone | matt | conservatives | bolster | davis |
| newt | larry | likewise | alert | blah |
| spur | light_skin_tone | spill | greenhouse | compliance |
| ❤️ | initiative | competitor | charles | jon |
| karl | elephant | atlanta | monitor | revoke |
| airline | dissolve | carlson | businessinsider_url | dominion |
| captain | tldr | initiate | furthermore | tobacco |
| mission | drilling | upgrade | evidently | jfk |
| port | andlt | steel | taylor | hurricane |
| rescue | joy | hunter | offshore | aids |
| rove | mar | reddit_url | wire | bell |
| cough | workplace | abusive | supposedly | ma |

Table 6: 100 words with highest aligned cosine distance

**Highest-frequent mistranslations**

| 0-20 | 20-40 | 40-60 | 60-80 | 80-100 |
|---|---|---|---|---|
| republican | 🤣 | michigan | permit | warming |
| democrats | mitch | johnson | nyc | o |
| republicans | ted | wisconsin | lefty | um |
| liberal | rand | ha | stupidity | eh |
| cruz | ah | nancy | suppression | vaccination |
| leftist | yea | mitt | pete | ga |
| unfortunately | lmao | korea | audit | mission |
| subreddit | hundred | youtube_url | kkk | depression |
| sander | ron | btw | charity | carson |
| warren | brother | beto | graham | ben |
| 😂 | everybody | centrist | woke | bc |
| kasich | kamala | pa | imgur_url | cuomo |
| yep | al | anyways | thomas | supposedly |
| bs | reddit_url | chamber | govt | rnc |
| harris | ohio | toward | pennsylvania | washingtonpost_url |
| desantis | communism | rush | garland | hrc |
| orange | got | found | tucker | san |
| rubio | maga | hunter | pipeline | sexist |
| approve | haha | coward | bigot | rioter |
| son | yup | alien | la | alabama |

Table 7: 100 highest-frequent mistranslating words

**Highest sense separation**

| 0-20 | 20-40 | 40-60 | 60-80 | 80-100 |
|---|---|---|---|---|
| r | cruz | gun | evangelical | goldwater |
| biden | obese | dem | we | ron |
| comment | fox | rand | desantis | stack |
| democrats | broadband | wedding | democratic | racist |
| republican | newt | climate | aoc | fascist |
| liberal | her | oxygen | absentee | suppression |
| gop | gingrich | paul | dems | devos |
| left | pete | abrams | install | president |
| trump | spur | white | fraud | incumbent |
| she | joe | civility | mayor | homosexual |
| he | supporter | receipt | party | overcome |
| conservative | conventional | right | student | neutrality |
| hillary | greenhouse | orange | sander | mature |
| clinton | theocracy | illegal | kamala | rubio |
| republicans | castro | salon | warren | min |
| democrat | medium | weaponize | competitive | federal |
| race | obama | newsmax | vote | wi |
| sanders | his | libertarian | establishment | end |
| pbs | bernie | hunter | hiv | majority |
| progressive | leftist | senate | transgender | ballot |

Table 8: 100 words with highest sense separation

# Bibliograhy